\pdfoutput=1
\documentclass[10pt, conference, compsocconf]{IEEEtran}

\let\labelindent\relax

\usepackage{graphicx}
\usepackage{balance}  % for  \balance command ON LAST PAGE  (only there!)
\usepackage[ruled,vlined]{algorithm2e}
\usepackage{algorithmic}
\usepackage{listings}
\usepackage[subrefformat=parens]{subcaption}
\usepackage{amssymb}
\usepackage{mathtools}

\makeatletter
\newcounter{code}

\newenvironment{code*}[1][htb]
  {% Update algorithm name
   \let\c@algocf\c@code% Update algorithm counter
   \begin{algorithm*}[#1]%
  }{\end{algorithm*}}
\DeclareRobustCommand*\cal{\@fontswitch\relax\mathcal}
\makeatother

\newcommand{\dFAB}{\texttt{dFAB\@}}
\newcommand{\sFAB}{\texttt{sFAB\@}}
\newcommand{\dFABimpl}{\dFAB\@}
\newcommand{\sFABimpl}{\sFAB\@}

\usepackage{url}

\usepackage{color}
\newcommand{\rc}{\color{red}}

\usepackage{enumitem}
\usepackage{comment}
\usepackage[normalem]{ulem}

\begin{document}

% ****************** TITLE ****************************************

\title{Distributed Bayesian Piecewise Sparse Linear Models}
%\title{A Practical Approach to Distributed Inference of Bayesian Piecewise Sparse Linear Models}

% ****************** AUTHORS **************************************

\author{
\IEEEauthorblockN{Masato Asahara}
\IEEEauthorblockA{NEC System Platform Research Laboratories\\
masahara@nec-labs.com}
\and
\IEEEauthorblockN{Ryohei Fujimaki}
\IEEEauthorblockA{NEC Data Science Research Laboratories\\
rfujimaki@nec-labs.com}
}

\maketitle

\begin{abstract}
The importance of interpretability of machine learning models has been increasing 
due to emerging enterprise predictive analytics, threat of data privacy, accountability of artificial intelligence in society, and so on.
Piecewise linear models have been actively studied to achieve both accuracy and interpretability.
They often produce competitive accuracy against state-of-the-art non-linear methods.
In addition, their representations (i.e., rule-based segmentation plus sparse linear formula) are often preferred by domain experts.
A disadvantage of such models, however, is high computational cost for simultaneous determinations of the number of ``pieces'' and cardinality of each linear predictor, 
which has restricted their applicability to middle-scale data sets.
This paper proposes a distributed factorized asymptotic Bayesian (FAB) inference of learning piece-wise sparse linear models on distributed memory architectures.
The distributed FAB inference solves the simultaneous model selection issue without communicating $O(N)$ data where $N$ is the number of training samples and achieves linear scale-out against the number of CPU cores.
%and in this context FAB/HMEs~(hierarchical mixture of experts using factorized asymptotic Bayesian inference) have recently addressed model selection issues of piecewise sparse linear models~(i.e., simultaneous determination of the number of regions, cardinality of each linear predictor and model parameters), as yet their applicability has been restricted to middle-scale data sets due to their high computational costs.
%This paper proposes a distributed FAB/HME algorithm, denoted as \dFAB, on distributed memory architectures. 
%Key advantages of \dFAB\ include: 1) small communication cost: \dFAB\ solves the simultaneous model selection issue without communicating $O(N)$ where $N$ is the number of training samples, and 2) small synchronization overhead: a distributed model selection algorithm in \dFAB\ naturally equalizes loads on multiple CPU cores, resulting in small synchronization overhead and high CPU usage.
Experimental results demonstrate that the distributed FAB inference achieves high prediction accuracy and performance scalability with both synthetic and benchmark data.
\end{abstract}

%\keywords{Machine learning; Distributed computing and algorithm; Interpretable models; Sparse models}

\section{Introduction}\label{sec:intro}
The importance of interpretability and transparency of machine learning models has been increasing 
due to emerging enterprise predictive analytics, threat of data privacy, accountability of artificial intelligence in society, and so on.
In data mining and machine learning academic community, the workshop named FAT/ML\footnote{\url{http://www.fatml.org/}} (fairness, accountability and transparency in machine learning) has been held every year since 2014. 
This momentum has grown by incorporating legal aspects of machine learning agents~(e.g., Symposium on Machine Learning and the Law in NIPS2016\footnote{\url{http://www.mlandthelaw.org/}}).
From government point of view, European Union enforces GDPR\footnote{\url{http://www.eugdpr.org/}} (general data protection regulation) which requires that the consequences of profiling (i.e., how models profile individuals) should be informed to the data subject.
%A Japanese ministry proposes basic rules of AI development at the G7 ICT Ministers' meeting in 2016, and the rules include transparency, interpretability and accountability of AI agents.
On the other hand, interpretable models restrict model representations, and the balance between interpretability and accuracy has been important research topics for decades~\cite{Freitas:2014:CCM:2594473.2594475}.

%In this context, transparent models such as sparse linear models or tree models are widely used despite the dramatic evolution of machine learning approaches, such as kernel machines~\cite{Vapnik:1995:NSL:211359}, boosting~\cite{Friedman00greedyfunction}, random forests~\cite{Breiman:2001:RF:570181.570182}, and deep neural networks~\cite{Hinton:2006}.

Piecewise linear models have been actively studied to achieve both accuracy and interpretability, which 
include from classical tree~\cite{cart84} or linear~\cite{Tibshirani94regressionshrinkage} ones to 
more advanced ones~\cite{Jordan94,NIPS2012_4725,OiwaF14,NIPS2012_4791}.
They often produce competitive accuracy against state-of-the-art non-linear methods on real-world datasets.
In addition, their representations (i.e., rule-based segmentation plus sparse linear formula) are often preferred by domain experts.
%Region-specific linear models have been actively studied (see \cite{Freitas:2014:CCM:2594473.2594475} for a comprehensive review of this literature).
%Such models include from classical ones such as decision trees~\cite{cart84,Murthy94asystem} and Lasso~\cite{Tibshirani94regressionshrinkage} to more advanced models such as hierarchical mixture of experts~(HME)~\cite{Jordan94}, Bayesian treed linear models~\cite{BTLMs}, local supervised learning through space partitioning~\cite{NIPS2012_4725}, partition-wise linear models~\cite{DBLP:conf/nips/OiwaF14}, informative projection ensembles~\cite{NIPS2012_4791} and so on.
%These models are non-linear and achieves high prediction accuracy but still interpretability due to their predictors are (sparse) linear in local coordinates.
%One of key challenges of learning piece-wise linear models is model selection. 
To the best use of these models, i.e., simple and accurate, we have to simultaneously determine the number of ``pieces'' and cardinality of each predictor.
However, such simultaneous model selection is essentially much more computationally demanding which has restricted their applicability to middle-scale data sets.

%Further, for analyzing Big Data, the size of a feature matrix easily exceeds memory capacity in a single computation node, and these motivate us to develop a distributed learning algorithm for piece-wise linear models.

%FAB/HMEs~\cite{fab_eto14,Wang:2015:TIA:2783258.2783407}~(HMEs using factorized asymptotic Bayesian inference~\cite{fujimaki12b,fujimaki2012factorized2}) have addressed this challenge by taking advantages of FAB inference which is recently developed Bayesian model selection and successfully applied many latent variable models~\cite{NIPS2013_5171,hayashi2015rebuilding,ICML-2015-LiuFFM}.
%Although FAB/HMEs efficiently solve the model selection in a single optimization path\footnote{FAB/HMEs do not require outer-loop for model selection but model is automatically adjusted in its EM-like alternative procedure.},  such simultaneous model selection is essentially much more computationally demanding, and the applicability of FAB/HMEs is yet restricted to middle-scale datasets.
%Further, for analyzing Big Data, size of a feature matrix easily exceeds memory capacity in a single computation node, and these strongly motivate us to develop a distributed learning algorithm for FAB/HMEs.

Meanwhile, for analyzing very large scale data (a.k.a. Big Data), 
the size of a feature matrix easily exceeds memory capacity in a single computation node.
Recent trends in distributed computational platforms for large-scale machine learning have been shifting 
from those based on distributed file systems~(e.g., Hadoop~\cite{hadoop}) to those based on distributed 
memories~(e.g., Spark~\cite{spark} and Parameter server~\cite{pm_osdi}).
%because many \emph{iterative} machine learning algorithms reuse intermediate results across multiple computations.
While Hadoop incurs substantial overhead due to the load of intermediate data from disks between computations, 
distributed memory architectures are able to avoid the need for disk access by storing data in memory across the computations.
Notably, Spark appears to be one of the most promising platforms for enterprise data analytics, and many distributed machine learning algorithms for Spark have recently been developed~\cite{sparkmllib,mli,graphx,alid,Lin14,Gopalani15,Qiu14,Dhar15,Moritz16,Kim16}.

This paper proposes a novel distributed algorithm for learning piecewise linear models on distributed memory architectures and an efficient implementation on Spark. Our contributions are summarized as below.

\noindent \textbf{Distributed Learning Algorithm:}
This paper develops a distributed learning algorithm of piecewise linear models with model selection.
Our technical contributions are mainly two-fold.
First, our algorithm linearly scales over the number of distributed workers and automates the model selection problem by taking advantages of recently-developed two techniques: 1) factorized asymptotic Bayesian hierarchical mixture of experts~(FAB/HME)~\cite{fab_eto14} 
for model selection of piecewise linear models and 2) median selection subset aggregation estimator~(MESSAGE)~\cite{NIPS2014_5328} for 
communication-efficient distributed feature selection.
Second, the MESSAGE algorithm independently processes data in each worker for communication efficiency and 
we observe this yields a bias in factorized information criterion (FIC)~\cite{fujimaki12b,fujimaki2012factorized2,NIPS2013_5171,hayashi2015rebuilding,ICML-2015-LiuFFM}, the model selection criterion of FAB/HME.
We derive an asymptotic correction term of this bias in FIC, which leads better feature selection of individual local models.

\noindent \textbf{Practical Design on Spark:}
This paper presents a design of our algorithm on Spark that helps to fully utilize distributed computation resources. 
%The major computational factor in the FAB/HME algorithm is on matrix computations which cannot be efficiently performed using Spark native APIs.
A resilient distributed dataset (RDD) is designed to perform iterative model optimization without shuffling data.
Further, we show that the sample-wise parallelization of our algorithm uses CPU resources much more efficiently than expert-wise one.
Our experimental results demonstrate that our algorithm and design achieves both high prediction accuracy and high scalability with both synthetic and benchmark data compared to state-of-the-art Spark machine learning libraries.

\section{Related Work}\label{sec:relatedwork}

%Work related to \dFAB\ can be classified into two main areas: distributed machine learning algorithms and distributed computing platforms for machine learning.
%
%\paragraph*{Distributed machine learning algorithms}

%At the beginning of the Big Data analytics era, several studies proposed ways to address issues in analyzing Big Data on a single node with limited computation power and memory space.
%For example, VFDT~\cite{Domingos:2000:MHD:347090.347107} is an online model learning system using Hoeffding trees which allow learning with a very short constant time per example, and it has strong guarantees of high asymptotic similarity to the corresponding batch trees.

Piecewise linear models have actively been studied to achieve both interpretability and accuracy.
Such models include from classical ones such as decision trees~\cite{cart84,Murthy94asystem} and Lasso~\cite{Tibshirani94regressionshrinkage} to more advanced models such as hierarchical mixture of experts~(HME)~\cite{Jordan94}, Bayesian treed linear models~\cite{BTLMs}, local supervised learning through space partitioning~\cite{NIPS2012_4725}, informative projection ensembles~\cite{NIPS2012_4791}, supersparse linear integer models~\cite{Ustun13} and so on.
Optimization of piecewise linear models is usually non-convex due to simultaneous optimizations of partitions and local models. 
Partition-wise linear models~\cite{OiwaF14} addressed this issue by formulating it as a structured-sparsity problem.
FAB/HMEs \cite{fab_eto14} induce sparseness both on tree structures and cardinalities of local models, and 
fully automate simultaneous model selection for learning piecewise linear models via FAB inference~\cite{fujimaki12b,hayashi2015rebuilding}.
Jialei et al.~\cite{Wang:2015:TIA:2783258.2783407} extended FAB/HMEs and incorporated non-linearity in local predictors to gain better accuracy by keeping a certain level of interpretability.
Ribeiro et al.~\cite{Ribeiro16} proposed to locally approximate non-linear models by linear models for model agnostic interpretability. 
As far as we know, the applications of sophisticated piecewise linear models have been limited to middle scale datasets due to their high computational costs.
%and to the best of our knowledge there have been no previous work on distributed learning of piecewise linear models.

Meanwhile, Spark~\cite{spark} appears to be one of the most promising platforms for distributed machine learning algorithms.
%The distributed memory-based computing architecture of Spark is quite suitable for distributed machine learning algorithms because, unlike disk-based architecture such as Hadoop, it avoids I/O overhead of iterative data accesses during the algorithm execution.
There are a large quantity of researches to realize distributed machine learning algorithms on Spark such as logistic regression~\cite{Lin14}, SVM~\cite{Lin14}, K-Means~\cite{Gopalani15}, LDA~\cite{Qiu14}, ADMM~\cite{Dhar15}, dominant cluster detection~\cite{alid}, graph algorithms~\cite{graphx} and so on.
Because of its high scale computing power, automation of hyper-parameter search on Spark such as \cite{Sparks15} is also an active research field.
Furthermore, Spark has gotten a lot of attention as a platform of deep learning recently~\cite{Moritz16, Kim16}.
These research outcomes are continuously integrated with Spark as its machine learning library called MLlib~\cite{mli,sparkmllib}.
As growing the proposals of cutting-edge technologies, application field of machine learning on Spark is spreading to industrial area including electric power~\cite{Zheng14}, telecommunication~\cite{Alsheikh16} and drug discovery~\cite{Harnie15}.
In this movement, the importance of Spark as a data science platform for KDD community is also growing as we can find several tutorials held in the last several KDD conferences~\cite{Shanahan15, Armbrust16, Agosta16}.
%Spark still continues to be improved in order to contribute advanced data mining and analytics like graph processing~\cite{Low12} and stream processing~\cite{Zaharia13}.
%Despite the improvement of Spark, it is a non-trivial research challenge to design distributed algorithms on distributed memory architectures to achieve high computing performance because of the complexity of the architecture~\cite{Dunner16}.
Despite increasing attentions of Spark-based high-scale machine learning, to the best of our knowledge, there are little studies on distributed learning of piecewise linear models (on Spark).

\section{Preliminary}\label{sec:sFAB}
%This section summarizes the original FAB/HMEs~\cite{fab_eto14} as a foundation for \dFAB.

\begin{figure}
\centering
\includegraphics[width=.6\linewidth]{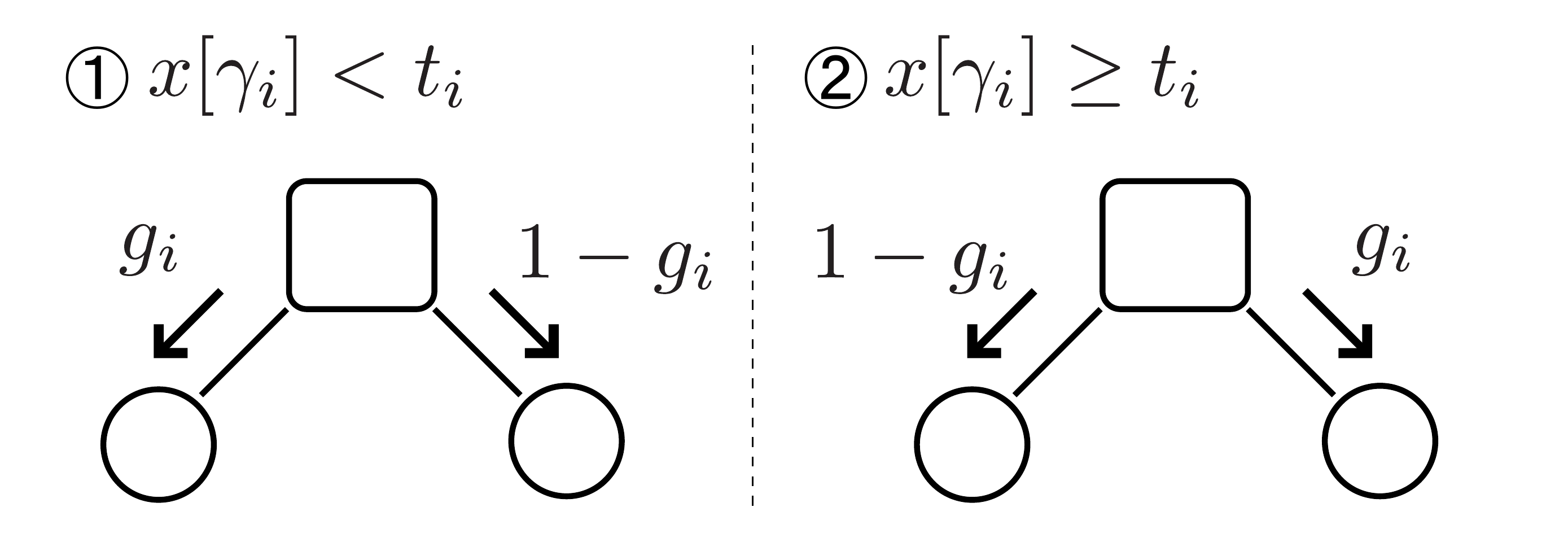}
\caption{Example of Bernoulli Gating Node.}
\label{fig:bgn}
\end{figure}

%\subsection{Piecewise Sparse Linear Models}
Probabilistic piecewise sparse linear models, or FAB/HME models~\cite{fab_eto14}, partition feature spaces using {\it gating functions} and assign a sparse expert in each partition. 
%Fig.~\ref{fig:hme} illustrates an example of HME models: 
%Given a sample $x$ the gating functions select an appropriate expert node for prediction, in a soft decision tree manner.
%\begin{figure}
%\centering
%\includegraphics[width=.45\linewidth]{figs/hme_structure}
%\caption{Example of HME model.}
%\label{fig:hme}
%\end{figure}
%
The FAB/HME model employs the Bernoulli gating function as follows:
\begin{equation}
g(x, \beta_i) := g_i U(t_i - x[\gamma_i]) + (1-g_i) U(x[\gamma_i] - t_i),
\end{equation}
where $\beta_i = (g_i, t_i, \gamma_i)$, $g_i \in [0,1]$, $U$ is a step function, 
$\gamma_i$ is the index w.r.t. the element of $x \in \mathbb{R}^D$, where $D$ is feature dimensionality,
and $t_i \in \mathbb{R}$ is an arbitrary value.
For example, when $x[\gamma_i] < t_i$, $g(x, \beta_i)=g_i$ and otherwise $g(x,\beta_i)=1-g_i$, as shown in Fig.~\ref{fig:bgn}.

Formally, the probabilistic models are defined as follows:
\begin{align}
p(y|x, \theta) = \sum_{j=1}^E \prod_{i \in \mathcal{E}_j} \psi_g(x,i,j) p(y|x, \phi_j),
\end{align}
where $x \in \mathbb{R}^D$ is an observation variable, $y$ is a target variable~($y \in \mathbb{R}$ for regression or $y \in \{1, -1\}$ for classification), $\theta = (\beta_1, \ldots, \beta_G, \\\phi_1, \ldots, \phi_E)$ represents model parameters, $E$ and $G$ are the numbers of experts and gating functions, respectively.
$\mathcal{E}_j (j=1, \ldots, E)$ denotes the $j$-th expert index set and contains all indices of the gating nodes on the unique path from the root node to the $j$-th expert node.
$\psi_g^{(i,j)}(x) := \psi(g(x, \beta_i), i, j)$ is the probability on the $i$-th gate, and $\prod_{i \in \mathcal{E}_j} \psi_g^{(i,j)}(x)$ expresses the probability of the $j$-th path, where $\psi(a, i, j) = a$ if the $j$-th expert is on the left sub-tree of the $i$-th gate, and $1-a$ otherwise.
%This paper employs both linear regression and logistic regression as models for experts.
The experts can be either linear regression, i.e., $p(y|x,\phi_{j}) = {\cal N}(y|w^{T}_{j}x,\sigma^{2}_{j})$ where $\phi_j = (w_j, \sigma^2_j)$, or linear logistic regression, i.e., $p(y|x, \phi_j) = 1 / (1 + \exp (- y \phi_j^T x))$.

The latent variable related to the $j$-th path is defined by $\zeta_j$, where $\zeta_j=1$ if $y|x$ is generated by the $j$-th node through the $j$-th path, and $\zeta_j=0$ otherwise.
The complete likelihood is defined, then, as follows:
\begin{align}
p(y^N|\zeta^N, x^N, \phi) &= \prod_{n=1}^N \prod_{j=1}^E p(y^{(n)}|x^{(n)}, \phi_j)^{\zeta_j^{(n)}}, \\
p(\zeta^N|x^N, \beta) &= \prod_{n=1}^N \prod_{j=1}^E \prod_{i \in \mathcal{E}_j} \psi_g^{(i,j)}(x^{(n)})^{\zeta_j^{(n)}},
\end{align}
where $\phi = (\phi_1, \ldots, \phi_E)$ and $\beta=(\beta_1, \ldots, \beta_G)$.
Prediction is executed by the $j_*$-th expert that maximizes the gating probability as follows:
\begin{align}
j_* = \arg\max_j \prod_{i \in \mathcal{E}_j} \psi_g^{(i,j)}(x).
\end{align}

The FAB/HME algorithm finds the best parameters and models by maximizing the \textit{factorized information criterion (FIC)}, which is an asymptotically accurate approximation of Bayesian marginal log-likelihood~\cite{fujimaki12b,hayashi2015rebuilding}, derived as follows:
\begin{align}
&FIC(y^N,x^N) = \max_q \Bigl\{ \mathbb{E}_q \Bigl[ \log p(y^N, \zeta^N|x^N, \theta) \uwave{- \sum_{i=1}^G \Bigl(} \label{eq:fic} \nonumber  \\
&\uwave{\frac{D_{\beta_i}}{2} \log \sum_{n=1}^N \sum_{j \in \mathcal{G}_i} q(\zeta_j^{(n)}) \Bigr) - \sum_{j=1}^E \frac{D_{\phi_j}}{2} \log \sum_{n=1}^N \ell_{j}^{(n)}q(\zeta_j^{(n)})} \Bigr] \nonumber \\
& + H_q \Bigr\}, 
\end{align}
where $q$ is any distribution on $\zeta^N$, $H_q$ is the entropy of $q$, and $\mathcal{G}_i$ is the $i$-th gating index set and contains all indices of the experts on the sub-tree of the $i$-th gating node.
$\ell_{j}^{(n)}$ is a scaling factor where $\ell_{j}^{(n)} = 1/\sigma^2_j$ for linear regression or $\ell_j^{(n)} = \mu_j^{(n)}(1 - \mu_j^{(n)})$ where $\mu_j^{(n)} = 1/(1 + \exp(-y^{(n)} \phi_j^T x^{(n)}))$ for logistic regression.
This optimization is conducted by alternating optimizations of $q$ and $\theta$ like expectation-–maximization algorithm.
Note that $D_{\phi_j}$ is equivalent to the L$_0$-norm of $\phi_j$, which induces sparsity of the model.
\cite{fab_eto14} has applied the forward-backward greedy~(FoBa) algorithm~\cite{foba} for least square experts, which has the tightest error 
bounds among state-of-the-art methods. %, such as Lasso~\cite{Tibshirani94regressionshrinkage}~(relax L$_0$ norm to L$_1$ norm) and orthogonal matching pursuit~\cite{omp}.
We can easily extend this idea to logistic regression by applying the gradient FoBa algorithm~\cite{liu14}, which has the same theoretical error bounds for 
general smooth convex functions.

\section{Distributed FAB/HME Algorithm}\label{sec:dFAB}
%\subsection{Problem Analysis}\label{subsec:problem}
This paper considers situations in which data size $N$ is much larger than data dimensionality $D$.
The designation of the distributed FAB/HME algorithm and architecture involves the following three challenges.
First, due to memory capacity limitation, no single worker node can load entire data, target, and variational distributions.
This prohibits the algorithm from employing a straightforward parallelization like that on a shared memory architecture.
The second challenge is to avoid huge communication overhead.
Particularly communications among worker nodes cause reallocation of data (a.k.a. data shuffling in Spark) in distributed memory architectures.
Third, for the ``map-reduce'' computational model which is one of the most popular modern distributed memory computation models, 
balancing or equalizing CPU loads is needed to minimize synchronization overhead.

Hereafter, we denote by $\mathcal{D}$ a dedicated server and by $w \in \mathcal{W}$ a worker node, where $\mathcal{W}$ is a set of worker nodes.
We assume that the training data are distributed on memories in $\forall w \in \mathcal{W}$\footnote{The training data is stored in a distributed file system and loaded on memory of workers in parallel.}.
The subscription $w$ denotes the $w$-th worker node.
We denote by $\mathcal{N}_w$ a sample index set of the $w$-th worker node.
In Algorithm~\ref{alg:fic} $\sim$ Algorithm~\ref{alg:comp}, $\star$ and $\bullet$ are executed on the worker nodes in parallel and the dedicated server in serial, respectively.

\subsection{Distributed FIC Computation}

%We redefine FIC by the lower bound of FIC as follows:
%\begin{align}
%FIC = \mathbb{E}_q [ \log p(y^N, \zeta^N | x^N, \theta)] - \sum_{i=1}^G \frac{D_{\beta_i}}{2} \log \sum_{n=1}^N \sum_{j \in \mathcal{G}_i} 	q(\zeta_j^{(n)}) - \sum_{j=1}^E \frac{D_{\phi_j}}{2} \log \sum_{n=1}^N q(\zeta_j^{(n)}) + H_q (q).
%\end{align}
The distributed computation of FIC, which is necessary for convergence determination, is described in Algorithm~\ref{alg:fic}.
The FIC calculation consists of two parts: 1)~the sum of expected log-likelihood and 2)~regularization.
The former requires $x^N$, $y^N$ and $q^{(t)}(\zeta^N)$, which are distributed on the worker nodes, 
and hence each worker node computes expected log-likelihoods of data in its memory, and then only its summation needs to be collected by the dedicated server, as shown in line~1 of Algorithm~\ref{alg:fic}.
This step requires only communication of scalar values between the dedicated server and individual worker nodes.
The latter is computed in the dedicated server, as shown in line~2 of Algorithm~\ref{alg:fic}.
\begin{algorithm}[t]
\caption{Distributed FIC Computation}
\label{alg:fic}
\begin{algorithmic}[1]
\REQUIRE $\bullet$ $N_{\beta_i}^{(t)}, N_{\phi_j}^{(t)}, \theta^{(t)}$
\REQUIRE $\star$ $q^{(t)}(\zeta^{(n)}), \theta^{(t)}$
\ENSURE $\bullet$ $FIC^{(t)}$
\STATE $\star$ $LL_w^{(t)} = \sum_{n \in \mathcal{N}_w} \Bigl( E_q [\log p(y^{(n)}, \zeta^{(n)} | x^{(n)}, \theta^{(t)})]  - q^{(t)}(\zeta^{(n)}) \log q^{(t)}(\zeta^{(n)})\Bigr)$ \quad for $\forall n \in \mathcal{N}_w$ and $\forall w \in \mathcal{W}$
\STATE $\bullet$ $FIC^{(t)} = \sum_{w=1}^W LL_w^{(t)} - \sum_{i=1}^G \frac{D_{\beta_i}^{(t)}}{2} \log N_{\beta_i}^{(t)} - \sum_{j=1}^E \frac{D_{\phi_j}^{(t)}}{2} \log N_{\phi_j}^{(t)}$
\end{algorithmic}
\end{algorithm}

\subsection{Distributed E-step}
The distributed computation of the E-step is described in Algorithm~\ref{alg:estep}.
First, $q(\zeta_{j}^{(n)})$ for $\forall n \in \mathcal{N}_{w}$ is calculated on $w \in \mathcal{W}$ as follows:
\begin{align}
\label{eq:fab_estep}
q^{(n,t)}_j := q^{(t)}(\zeta_j^{(n)}) &\propto \prod_{i \in \mathcal{E}_j} \psi_g^{(i,j,t-1)}(x^{(n)}) p(y^{(n)}|x^{(n)}, \phi_j^{(t-1)} )  \nonumber \\
&\uwave{\exp \Bigl\{ \sum_{i \in \mathcal{E}_j} \frac{-D_{\beta_i}}{2N_{\beta_i}^{(t-1)}} + \frac{-D_{\phi_j} \ell_j^{(n,t-1)}}{2N_{\phi_{j}}^{(t-1)}} \Bigr\}},
\end{align}
The expected numbers of samples on experts and gates, denoted by $N_{\phi_j}^w$ and $N_{\beta_i}^w$, 
are then collected by the dedicated server, and, therefore, only two scalar values are communicated.
In line~3, $N_{\phi_j}^{(t)}$ and $N_{\beta_i}^{(t)}$ are computed and globally shared on the dedicated server.
%$N_{\phi_j}^{(t)}$ and $N_{\beta_i}^{(t)}$ are distributed to the worker nodes for subsequent steps, as shown in the output of Algorithm~\ref{alg:estep}.
It is known that exponentiated regularization~(the waved part of \eqref{eq:fab_estep}) 
derived from FIC~(the waved part of \eqref{eq:fic}) eliminates redundant latent variables through EM iterations~\cite{fujimaki12b}.
Such ``non-effective'' experts are eliminated from the model as follows:
\begin{align}
q^{(n,t)}_j = 0 \quad \mbox{if $ N_{\phi_j}^{(t)} < \delta$ otherwise} \quad q^{(n,t)}_j/Q_j^{(t)} , \label{eq:shrink}
\end{align}
where $\delta$ and $Q_j^{(t)}$ are a threshold value and a normalization constant for $\sum_{j=1}^{E} q^{(n,t)}_j=1$. 
This {\it shrinkage} process automatically determines piecewise space partitioning structures in a Bayesian-principled fashion.
\begin{algorithm}[t]
\caption{Distributed E-step}
\label{alg:estep}
\begin{algorithmic}[1]
\REQUIRE $\star$ $N_{\beta_i}^{(t-1)}, N_{\phi_j}^{(t-1)}, \theta^{(t-1)}$
\ENSURE $\bullet$ $N_{\phi_j}^{(t)}, N_{\beta_i}^{(t)}$
\ENSURE $\star$ $q^{(t)}(\zeta_j^{(n)}), N_{\phi_j}^{(t)}, N_{\beta_i}^{(t)}$
\STATE $\star$ Calculate $q^{(n,t)}_j$ using \eqref{eq:fab_estep} for $\forall n \in \mathcal{N}_w, \forall w \in \mathcal{W}$.
\STATE $\star$ $N_{\phi_j}^w = \sum_{n \in \mathcal{N}_w} q^{(n,t)}_j$ and $N_{\beta_i}^w = \sum_{j \in \mathcal{G}_i} N_{\phi_j}^w$
\STATE $\bullet$ $N_{\phi_j}^{(t)} = \sum_{w \in \mathcal{W}} N_{\phi_j}^w$ and $N_{\beta_i}^{(t)} = \sum_{w \in \mathcal{W}} N_{\beta_i}^w$
\STATE $\bullet$ Eliminate irrelevant experts and gates using \eqref{eq:shrink}.
\end{algorithmic}
\end{algorithm}

\subsection{Distributed M-step: Bernoulli Gates}
%As shown in Algorithm~\ref{alg: gate opt}, 
The gate optimization requires a set of 
split candidate points, i.e., $\mathcal{B}_d = \{t_i | \mbox{discretized }\mbox{domain } \mbox{of } x_d \}$, for each dimension $d$ of $x$. 
To consistently aggregate distributed calculations, all worker nodes must share $\mathcal{B}_d$ for $\forall d$.
For this purpose, $\mathcal{B}_d$ is computed at the beginning of the FAB-EM algorithm on the basis of Algorithm~\ref{alg:split}.
First, on each worker node, the maximum and minimum values of $x_d^{(n)}$ are computed and then collected by the dedicated server.
Here a vector of the size $D$ is transferred from each worker node to the dedicated server.
$\mathcal{B}_d$ is computed on the basis of lines 2 and 3 in Algorithm~\ref{alg:split}.
$\mathcal{B} = \{\mathcal{B}_d\}$ is distributed to all worker nodes. This process communicates a matrix of the size $DT_{\max}$.
\begin{algorithm}[t]
\caption{Split Points Calculation}
\label{alg:split}
\begin{algorithmic}[1]
\REQUIRE $\bullet$ $T_{\max}$: number of the maximum split points
\ENSURE $\star$ $\mathcal{T}_d$ for $\forall d$.
\STATE $\star$ Calculate $x_{dw}^{\max} = \max_{n \in \mathcal{N}_w} \{ x^{(n)}_d \}$ and $x_{dw}^{\min} = \min_{n \in \mathcal{N}_w} \{x^{(n)}_d \}$ for $\forall d$ and $\forall w$.
\STATE $\bullet$ Calculate $x_d^{\max} = \max_w \{ x_{dw}^{\max} \}$ and $x^{\min}_d = \max_w \{ x_{dw}^{\min} \}$.
\STATE $\bullet$ Calculate $\mathcal{T}_d$ by splitting the range $[x_d^{\min}, x_d^{\max}]$ into $T_{\max}$ bins with the equal width.
\end{algorithmic}
\end{algorithm}

The distributed gate optimization is summarized in Algorithm~\ref{alg:bern}.
First, some intermediate statistics are computed in a distributed manner as follows~(at line~1-3):
\begin{align}
\rho^L_w (\gamma_i, t_i) = \sum_{n \in \mathcal{N}_w \cap \mathcal{T}_{si}} \sum_{j \in \mathcal{G}_{iL}} q^{(n,t)}_j \label{eq: rhoL}\\
\rho^R_w (\gamma_i, t_i) = \sum_{n \in \mathcal{N}_w \cap \mathcal{T}_{li}} \sum_{j \in \mathcal{G}_{iR}} q^{(n,t)}_j \label{eq: rhoR}
\end{align}
where $\mathcal{T}_{li}, \mathcal{T}_{si}$ are the sets of samples whose $\gamma_i$-th dimension is larger or smaller than $t_i$,
$\mathcal{G}_{iL}$ contains all indices of the expert nodes on the left sub-tree of the $i$-th gating node, and 
$\mathcal{G}_{iR}$ is similarly defined for the right sub-tree of the $i$-th gating node.
The two matrices w.r.t. $\rho^L_w$ and $\rho^R_w$, each of which has $DT_{\max}$ elements, are collected by the dedicated server.
Then, on the dedicated server, the $i$-th gate parameter $\beta_i^{(t)}$ is computed as described in lines 4-8 in Algorithm~\ref{alg:bern}, and 
$\beta_i^{(t)}$, containing three scalar values, is distributed to all worker nodes.
%
%Note that $\mathcal{T}$ and $\mathcal{G}$ are both functions of $\gamma_i$ and $t_i$.
%
\begin{algorithm}[t]
\caption{Distributed Bernoulli Gate Optimization}
\label{alg:bern}
\begin{algorithmic}[1]
\REQUIRE $\star$ $\mathcal{T}_d, N_{\beta_i}^{(t)}, q^{(t)} (\zeta_j^{(n)})$
\ENSURE $\star$ $\beta_i^{(t)}$
\FOR{$\mathcal{T}_{\gamma_i}$ where $\gamma_i = 1,\ldots, D$}
%\FOR{$\mathcal{T}_{\gamma_i}$}
\STATE $\star$ Calculate $\rho^L_w (\gamma_i, t_i)$ and $\rho^R_w (\gamma_i, t_i)$ for $\forall w \in \mathcal{W}$ using \eqref{eq: rhoL} and \eqref{eq: rhoR}.
%\ENDFOR
\ENDFOR
\FOR{$\mathcal{T}_{\gamma_i}$ where $\gamma_i = 1,\ldots, D$}
%\FOR{$\mathcal{T}_{\gamma_i}$}
\STATE $\bullet$ Calculate $\rho^L (\gamma_i, t_i)  = \sum_{w \in \mathcal{W}} \rho_{w}^L (\gamma_i, t_i)$ and $\rho^R (\gamma_i, t_i) = \sum_{w \in \mathcal{W}} \rho_{w}^R (\gamma_i^{(t)}, t_i^{(t)})$.
\STATE $\bullet$ Calculate $\xi_i(\gamma_i, t_i)$ where $g_i (\gamma_i, t_i) = (\rho^L (\gamma_i, t_i) + \rho^{R} (\gamma_i, t_i)) / N_{\beta_i}^{(t)}$.
%\ENDFOR
\ENDFOR
\STATE $\bullet$ Calculate $\gamma_i^{(t)}, t_i^{(t)} = \arg\max_{\gamma_i, t_i} \xi_i (\gamma_i, t_i)$, and $g_i^{(t)} = g_i (\gamma_i^{(t)}, t_i^{(t)})$.
\end{algorithmic}
\end{algorithm}

\subsection{Distributed M-step: Sparse Experts}
For optimizing experts in the M-step, we have to distribute L$_0$ regularized optimization,
to which well-studied approaches using distributed gradient or proximal methods are not applicable.
We address this issue by applying a recently-developed median selection subset aggregation estimator~(MESSAGE) algorithm~\cite{NIPS2014_5328}.
%The MESSAGE algorithm first selects features by majority voting for feature selection results on subsets~(i.e., worker nodes) and 
%then estimates a sparse weight by aggregating estimators with only selected features on the subsets.
%The idea of the MESSAGE algorithm is quite simple, but it has strong theoretical guarantees, such as those for consistency of feature selection and weight estimation.

The distributed optimization of sparse experts is described in Algorithm~\ref{alg:comp}.
First, feature selection is performed using the FoBa algorithm on each worker node as follows~(line~1):
\begin{align}
\phi_{jw}^{(t+\frac{1}{2})} &= \arg\max_{\phi_j} |\mathcal{W}| \sum_{n \in \mathcal{N}_w} q^{(n,t)}_j  \log p(y^{(n)}|x^{(n)}, \phi_j) \label{eq: dist fs} \\
&- (|\mathcal{W}|-1)D_{\phi_j} - \frac{D_{\phi_j}}{2} \log \sum_{n=1}^N \ell_{j}^{(n, t-1)} q^{(n,t)}_j.  \nonumber
\end{align}
The detailed derivation of \eqref{eq: dist fs} is discussed in the next sub-section.
$\phi_{jw}^{(t+\frac{1}{2})}$ is then once collected by the dedicated server, and majority voting is performed to determine a feature set as follows~(line~2):
\begin{align}
F_{j} = \{ d \  | \ \sum_{w \in \mathcal{W}} [\phi_{jw}^{(t+\frac{1}{2})}]_d \ge \frac{|\mathcal{W}|}{2} \}. \label{eq: voting}
\end{align}
The feature set $F_{j}$ is then distributed to all workers, and parameter estimation is performed using only features in $F_j$ as follows~(line~3):
\begin{align}
\phi_{jw}^{(t)} = \arg\max_{\phi_j}& \sum_{n \in \mathcal{N}_w}  q^{(n,t)}_j \log p(y^{(n)}|x^{(n)}, \phi_j(F_j)), \label{eq: dist ls}
\end{align}
where $\phi_j(F_j)$ means that parameters not included in $F_j$ are fixed to zero.
Finally, $\phi_{jw}^{(t)}$ is again collected by the dedicated server, and the weight aggregation is performed to estimate the weight vector as follow~(line~4):
\begin{align}
\phi_{j}^{(t)} = \frac{1}{|\mathcal{W}|} \sum_{w \in \mathcal{W}} \phi_{jw}^{(t)}, \label{eq: agg}
\end{align}
where $\phi_{j}^{(t)}$ is distributed to all worker nodes.
%This process communicates vectors of the size $D$ four times~($\phi_{jw}^{(t+\frac{1}{2})}$, $F_{j}$, $\phi_{jw}^{(t)}$, and $\phi_{j}^{(t)}$).
Note that the majority voting and the weight aggregation are lightweight computations in comparison to the L0 optimization processes, and their executions on the dedicated server do not affect the parallel performance.
\begin{algorithm}[t]
\caption{Distributed Sparse Experts Optimization}
\label{alg:comp}
\begin{algorithmic}[1]
\REQUIRE $\star$ $q^{(t)} (\zeta^{(n)}_j), \phi_j^{(t-1)}$
\ENSURE $\star$ $\phi^{(t)}_j$
\STATE $\star$ Calculate $\phi_{jw}^{(t+\frac{1}{2})}$ using \eqref{eq: dist fs} for $\forall w \in \mathcal{W}$.
\STATE $\bullet$ Perform the majority voting using \eqref{eq: voting}.
\STATE $\star$ Estimate $\phi_{jw}^{(t)}$ using \eqref{eq: dist ls}.
\STATE $\bullet$ Compute $\phi_{j}^{(t)}$ by aggregating $\phi_{jw}^{(t)}$ using \eqref{eq: agg}.
\end{algorithmic}
\end{algorithm}

\subsection{Correction of FIC in M-step}
This subsection explains the derivation of \eqref{eq: dist fs}. In \eqref{eq:fic}, the terms related to expert optimization can be summarized as follows:
\begin{align}
\sum_{n=1}^N q^{(n,t)}_j  \log p(y^{(n)}|x^{(n)}, \phi_j) - \frac{D_{\phi_j}}{2} \log \sum_{n=1}^N \ell_{j}^{(n, t-1)} q^{(n,t)}_j. \label{eq: mfic}
\end{align}
The first (loss) and the second (regularizer) terms are $O(N)$ and $O(\log N)$, respectively. 
Therefore, in distributed M-step, if we simply replace $\sum_{n=1}^N$ by $\sum_{n \in \mathcal{N}_w}$, it changes the balance between the first and the second terms.
On the other hand, if we rescale the first term as follows:
\begin{align}
|\mathcal{W}| \sum_{n \in \mathcal{N}_w} q^{(n,t)}_j  \log p(y^{(n)}|x^{(n)}, \phi_j) - \frac{D_{\phi_j}}{2} \log \sum_{n=1}^N \ell_{j}^{(n, t-1)} q^{(n,t)}_j,
\end{align}
it causes different bias in FIC by reusing the first term~(i.e., the first term is evaluated $|\mathcal{W}|$ times on a single data set).

In order to avoid this bias, we consider an asymptotic approximation of the first term as follows\footnote{The derivation can be obtained in a similar manner of the derivation of Akaike's information criterion~\cite{aic}.}:
\begin{align}
&\sum_{n \in \mathcal{N}_w} q^{(n,t)}_j  (\log p(y^{(n)}|x^{(n)}, \phi_j^*) - \log p(y^{(n)}|x^{(n)}, \hat{\phi}_j)) \approx - D_{\phi_j} \nonumber  \\
&\sum_{n=1}^N q^{(n,t)}_j  (\log p(y^{(n)}|x^{(n)}, \phi_j^*) - \log p(y^{(n)}|x^{(n)}, \bar{\phi}_j)) \approx - D_{\phi_j} \label{eq: app}
\end{align}
where $\phi^*$ is the true parameter and $\hat{\phi}_j$ and $\bar{\phi}_j$ are the maximizer of $\sum_{n \in \mathcal{N}_w} q^{(n,t)}_j  \log p(y^{(n)}|x^{(n)}, \phi_j)$ and $\sum_{n=1}^N q^{(n,t)}_j  \log p(y^{(n)}|x^{(n)}, \phi_j)$, respectively.
By taking into account that $\mathcal{N}_w$ can be considered as a random subset of $\mathcal{N}$, we have:
\begin{align}
\sum_{n=1}^N q^{(n,t)}_j  &\log p(y^{(n)}|x^{(n)}, \phi_j) \label{eq: ave} \\
&= |\mathcal{W}| \mathbb{E}_{\mathcal{N}_w}[ \sum_{n \in \mathcal{N}_w} q^{(n,t)}_j  \log p(y^{(n)}|x^{(n)}, \phi_j^*)] \nonumber
\end{align}
where $E_{\mathcal{N}_w}$ is the expectation over the randomness on $\mathcal{N}_w$.
By combining \eqref{eq: mfic}, \eqref{eq: app} and \eqref{eq: ave}, we have \eqref{eq: dist fs}.

\section{Efficient Design on Spark}\label{sec:design}

This section describes an efficient design of the distributed FAB/HME algorithm on Spark.
Hereinafter we refer to the distributed FAB/HME as \dFABimpl\@ and the original FAB/HME as \sFABimpl\@.

\subsection{RDD Structure and Execution Flow}
A resilience distributed dataset~(RDD) is the base in Spark distributed computation, 
on whose elements all distributed computations on Spark are performed, i.e., Spark processes each element of an RDD in parallel.
An RDD is an immutable and partitioned collection of records and can only be created from persistent data or other RDDs via transformations.
A standard RDD design might assign one data instance to an element of the RDD.
However, major computations of FAB/HME algorithms rely on matrix computations, and the RDD for \dFABimpl\@ has to be designed to process them efficiently\footnote{Although we can use \texttt{mapPartition} function to collect a part of training data from a partition of the RDD, \texttt{mapPartition} passes an iterator of samples to the closure, which causes many iterator accesses and significant computational overhead.}.
For convenience, let us introduce a few notations. 
Let  $X_{e}$, $Y_{e}$, and $Q_{e}$ be matrices whose elements are $x^{(n)}$ for $\forall n \in \mathcal{N}_{e}$,
$y^{(n)}$ for $\forall n \in \mathcal{N}_{e}$, and $q(\zeta^{(n)})$ for $\forall n \in \mathcal{N}_{e}$, respectively, where $\mathcal{N}_{e} \subset \mathcal{N}_{w}$.
Note that each element of $Q_{e}^{(t)}$ is $q^{(t)}(\zeta^{(n)})$ for $\forall n \in \mathcal{N}_{e}$.
Also, $L_{e}$ is a $|\mathcal{N}_e| \times E$ matrix, and the $(n,j)$-th element of $L_{e}^{(t)}$ is:
\begin{equation*}
\log \prod_{i \in \mathcal{E}_j} \psi_g^{(i,j,t-1)}(x^{(n)}) p(y^{(n)}|x^{(n)}, \phi_j^{(t-1)} ).
\end{equation*}

%as the CPU profiling of \sFABimpl\ implies~(Fig.~\ref{fig:fab_proctime_ratio}), 
%71\% of computational time is consumed in M-step even with a sophisticated BLAS library and optimized codes in C++.
%Thus, it is essential to design the RDD so that those optimized serial libraries are utilized in \texttt{map} functions\footnote{Although we can use \texttt{mapPartition} function to collect a part of training data from a partition of the RDD, \texttt{mapPartition} passes an iterator of samples to the closure, which causes many iterator accesses and significant computational overhead.}.
%\begin{figure}[t]
%\centering
%\includegraphics[width=.7\linewidth]{figs/sFAB_proctime_ratio}
%\caption{Ratio of computational time of \sFAB.}
%\label{fig:fab_proctime_ratio}
%\end{figure}

\begin{figure}[t]
\centering
\includegraphics[width=.4\linewidth]{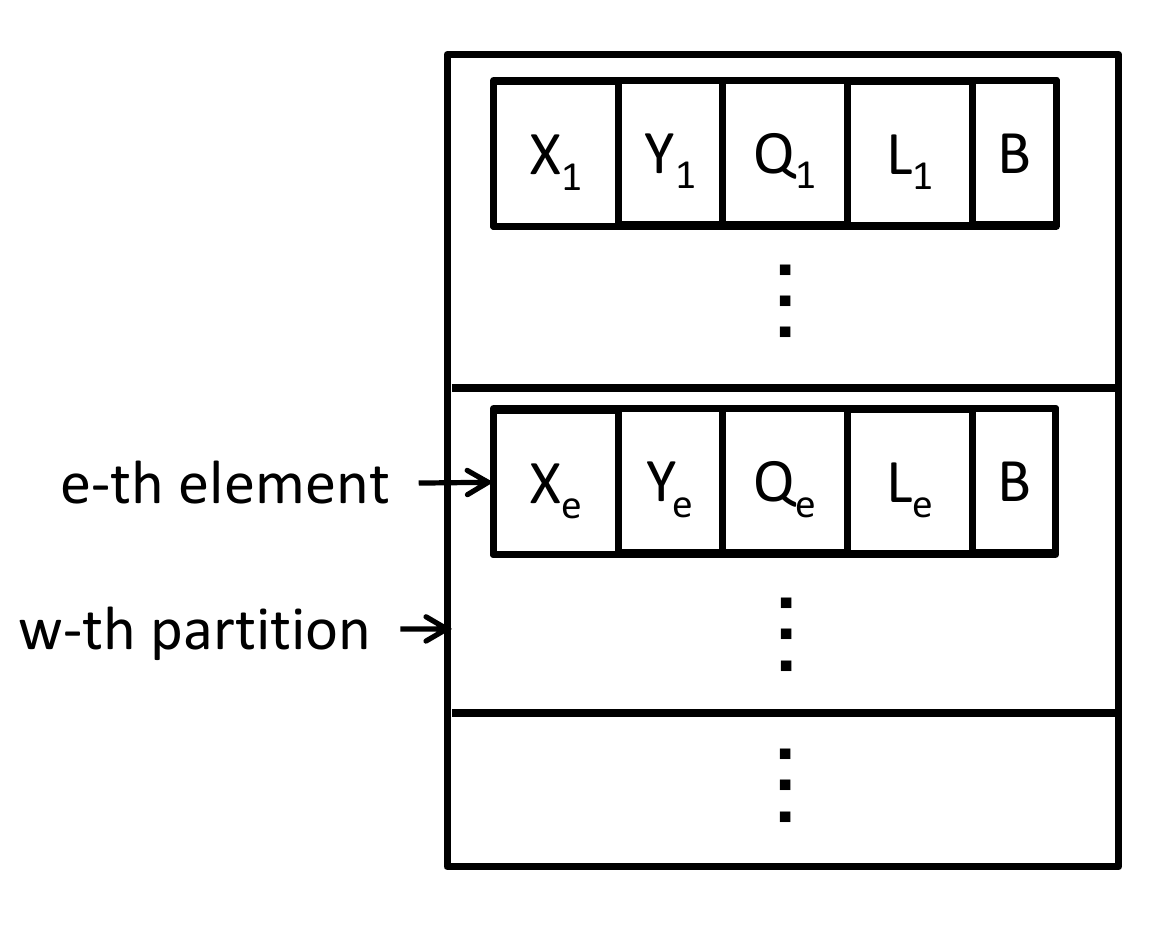}
\caption{RDD structure of \dFABimpl\@.}
\label{fig:RDD}
\end{figure}

\begin{figure}[t]
\centering
\includegraphics[width=.8\linewidth]{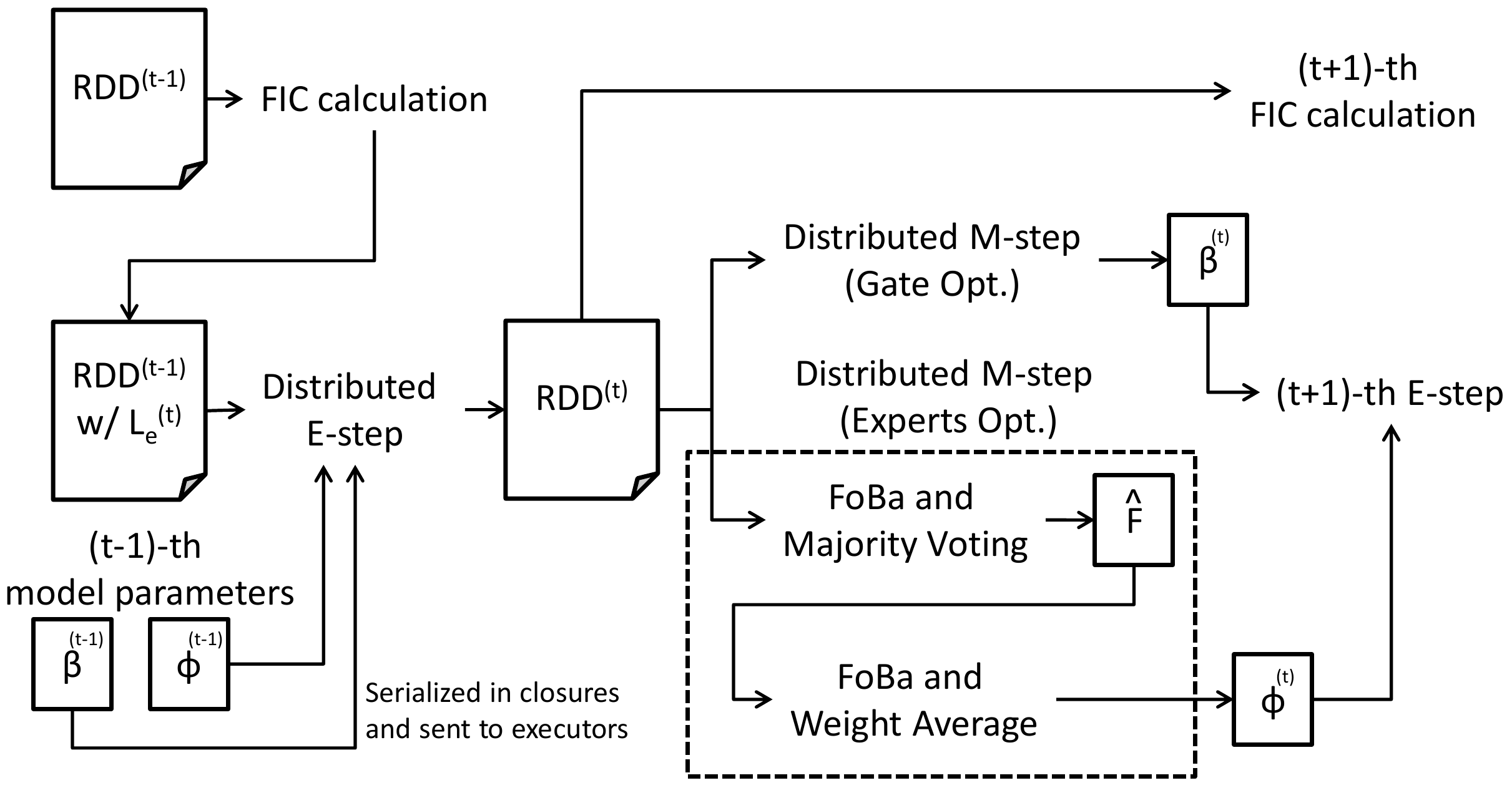}
\caption{Execution flow of \dFABimpl\@.}
\label{fig:execflow}
\end{figure}

Fig.~\ref{fig:RDD} illustrates the RDD structure in \dFABimpl.
One partition contains multiple elements that are data units to which map functions are applied, 
and one element of $w$-th partition consists of a tuple of $(X_{e}, Y_{e}, Q_{e}, L_{e}, \mathcal{B})$, where $e$ is the index of an element and $w$ corresponds to the worker node index $w$\footnote{For the evaluations in this paper we set the number of elements so that it equals the number of workers.}.%on the basis of empirical studies.}% issues such as increasing overheads of task serializations and deserializations.}.
%~(hence, the worker node index $w$ also expresses the $w$-th partition and element).
%$X_{e}$, $Y_{e}$, $Q_{e}$, $L_{e}$ and $\mathcal{B}$ are allocated once (and only once) 
%in a sequential space on memory.
%With this tuple element, the \dFABimpl\ is able to fully enjoy optimized libraries\footnote{The \dFABimpl\ utilizes Java Native Interface (JNI) to call C++ optimization libraries for gate optimization and FoBa, which are used in the \sFABimpl\@. The overhead of JNI is negligible in comparison to the computation cost of the serial optimization processes.}.
%It is worth noting that the best practice of RDD designs often suggests one partition to contain a set of elements and assign two or three tasks (i.e., partitions) to one CPU core~\cite{spark-tuning}.
%However, the \dFABimpl\ takes one-element-to-one-partition design because the M-step expert optimization is so heavy that the overhead of increasing number of elements, e.g., list scanning and closure serialization and deserialization, exceeds the improvement of CPU usage by increasing number of elements per partition.
%
Fig.~\ref{fig:execflow} illustrates the execution flow of \dFABimpl.
Let us denote the RDD after the $t$-th EM iteration by RDD$^{(t)}$, whose element is $(X_{e}, Y_{e}, Q_{e}^{(t)}, L_{e}^{(t)}, \mathcal{B})$.
\begin{itemize}[noitemsep,nolistsep,leftmargin=*]
\item In the $t$-th FIC calculation, the driver process invokes \texttt{map} transformation with a closure executing Step 1 of Algorithm~\ref{alg:fic}, which involves model parameters $\beta^{(t-1)}$ and $\phi^{(t-1)}$ as its parameters. Here, each executor computes $L_{e}^{(t)}$ as an intermediate outcome, and each RDD element is updated from $(X_{e}, Y_{e}, Q_{e}^{(t-1)}, L_{e}^{(t-1)}, \mathcal{B})$ to $(X_{e}, Y_{e}, Q_{e}^{(t-1)}, L_{e}^{(t)}, \mathcal{B})$. Note that, if the eliminated experts (Step 4 of Algorithm~\ref{alg:fic}) are identified in the driver process, $Q_{e}^{(t-1)}$s are updated in the executors by simply setting the corresponding column to be zero in order to avoid reallocation of the RDD itself.
\item In the $t$-th E-step, the driver process invokes a \texttt{map} transformation with a closure executing Steps 1 and 2 in the Algorithm~\ref{alg:estep}. %, which involves model parameters, $N_{\phi_{j}}^{(t-1)}$ and $N_{\beta_{i}}^{(t-1)}$, as its parameters.
Note that we avoid recalculation of expert-wise likelihood by storing $L_{e}^{(t)}$ as a part of the RDD element. Here, each RDD element is updated from $(X_{e}, Y_{e}, Q_{e}^{(t-1)}, L_{e}^{(t)}, \mathcal{B})$ to  $(X_{e}, Y_{e}, Q_{e}^{(t)}, L_{e}^{(t)}, \mathcal{B})$.
Through the FIC calculation and E-step, we obtain RDD$^{(t)}$ without re-allocating or shuffling any large scale data.
\item The $t$-th M-step is performed on the basis of Algorithm~\ref{alg:bern} and \ref{alg:comp}. This does not change RDD$^{(t)}$. %The next subsection discusses the M-step computation in details.
\end{itemize}

\subsection{Advantage of Sample-wise Parallelization over Expert-wise Parallelization} \label{sec:sample-wise-par}
\begin{figure}[t]
\begin{minipage}[b]{.45\linewidth}
\centering
\includegraphics[width=\linewidth]{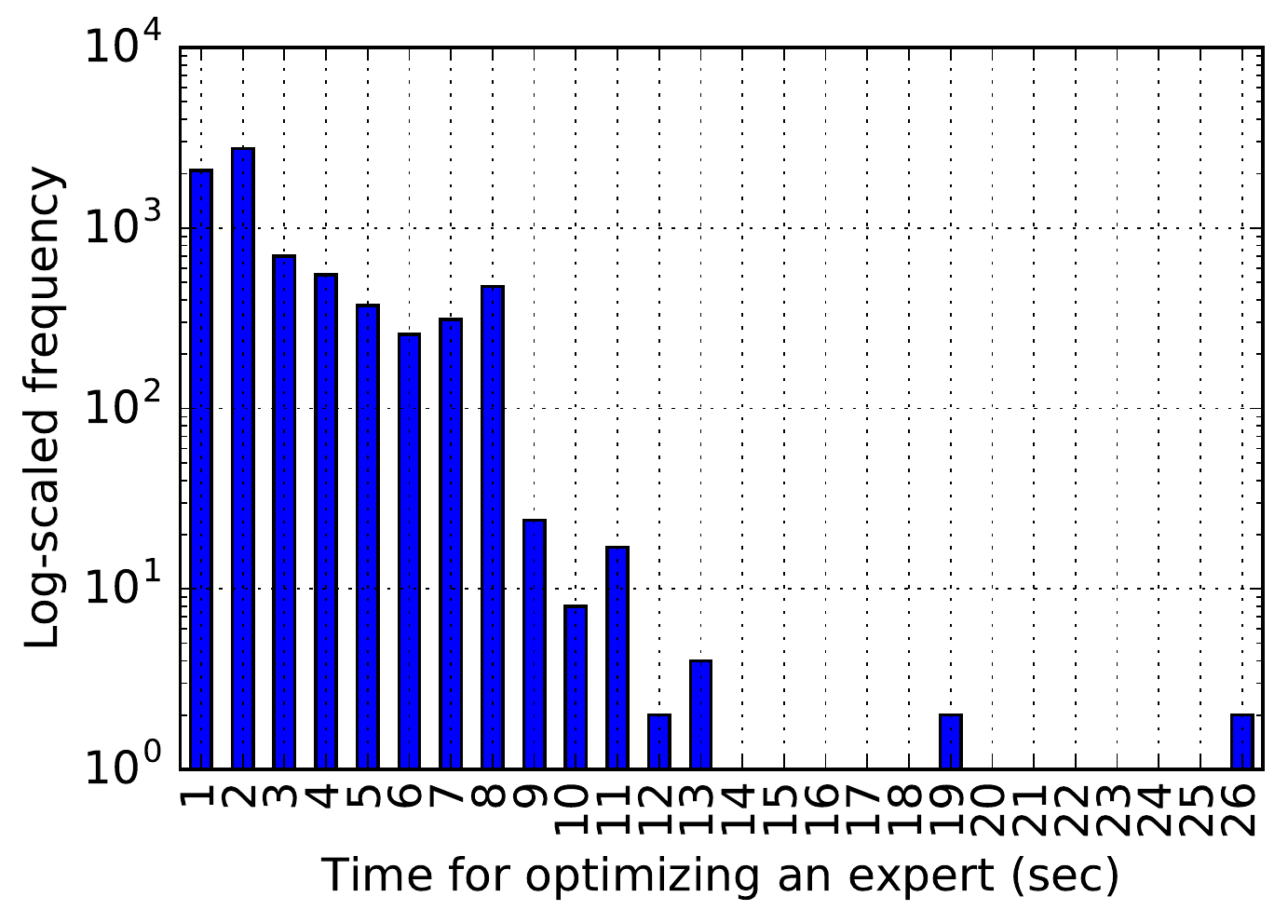}
\caption{Histogram of time for experts optimization.}
\label{fig:expert_opt_time}
\end{minipage}
\begin{minipage}[b]{.1\linewidth}
\end{minipage}
\begin{minipage}[b]{.45\linewidth}
\centering
\includegraphics[width=\linewidth]{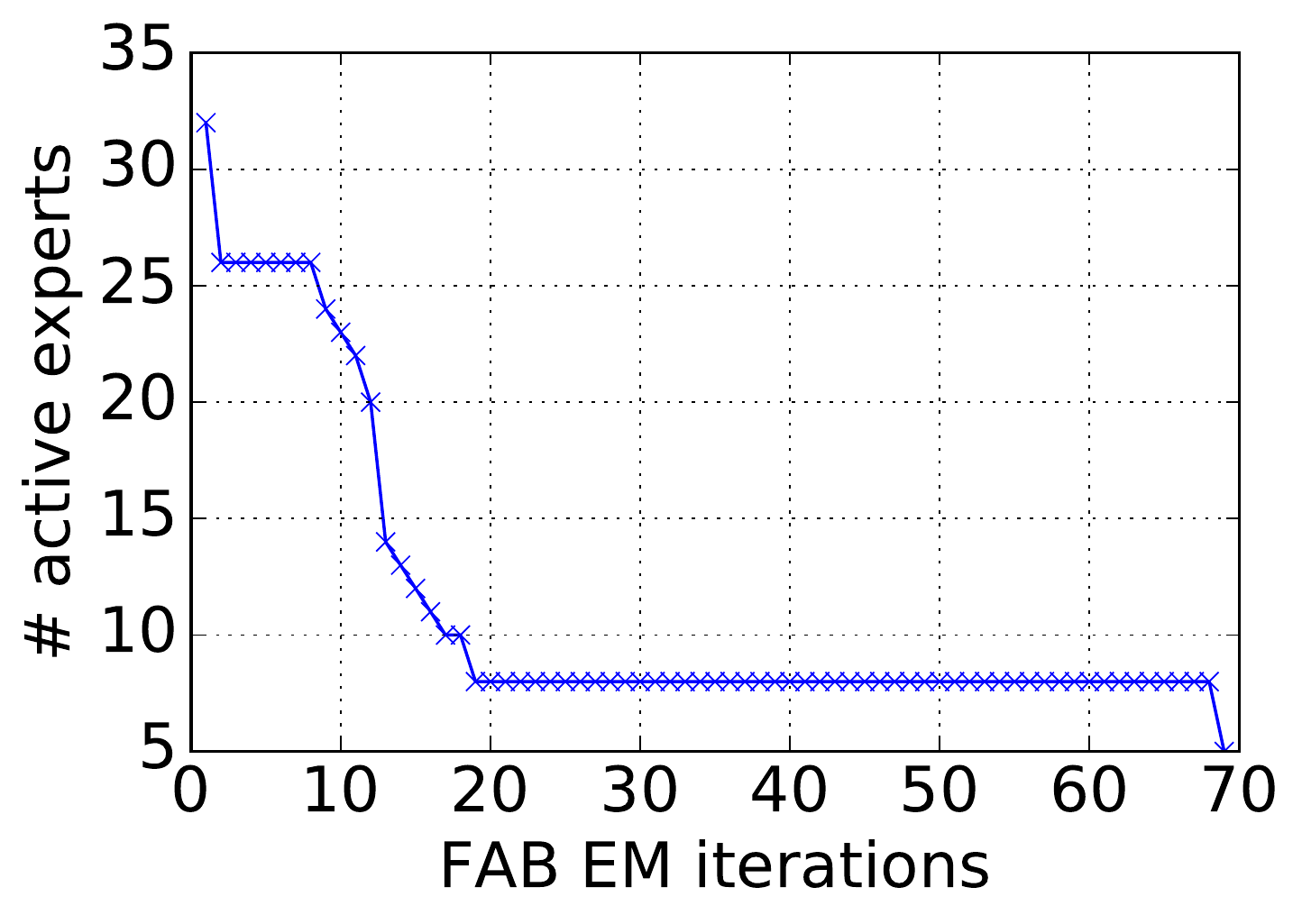}
\caption{Number of active experts over FAB EM iterations.}
\label{fig:history_num_experts}
\end{minipage}
\end{figure}

\begin{figure}[t]
\centering
\begin{minipage}[b]{.4\linewidth}
\centering
\includegraphics[keepaspectratio, width=\linewidth]{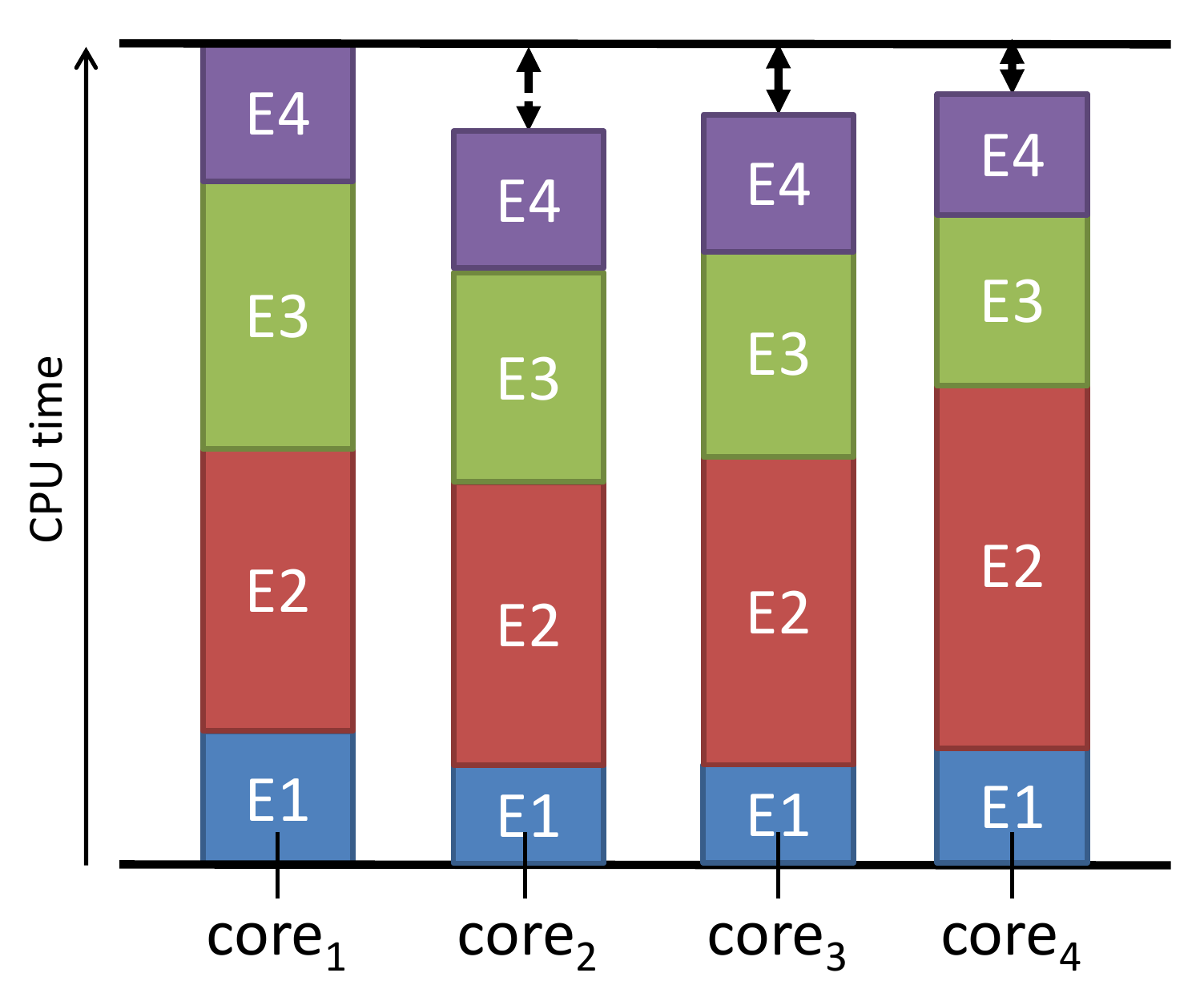}
\subcaption{Sample-wise}\label{fig:sample-wise}
\end{minipage}
\begin{minipage}[b]{.4\linewidth}
\centering
\includegraphics[keepaspectratio, width=\linewidth]{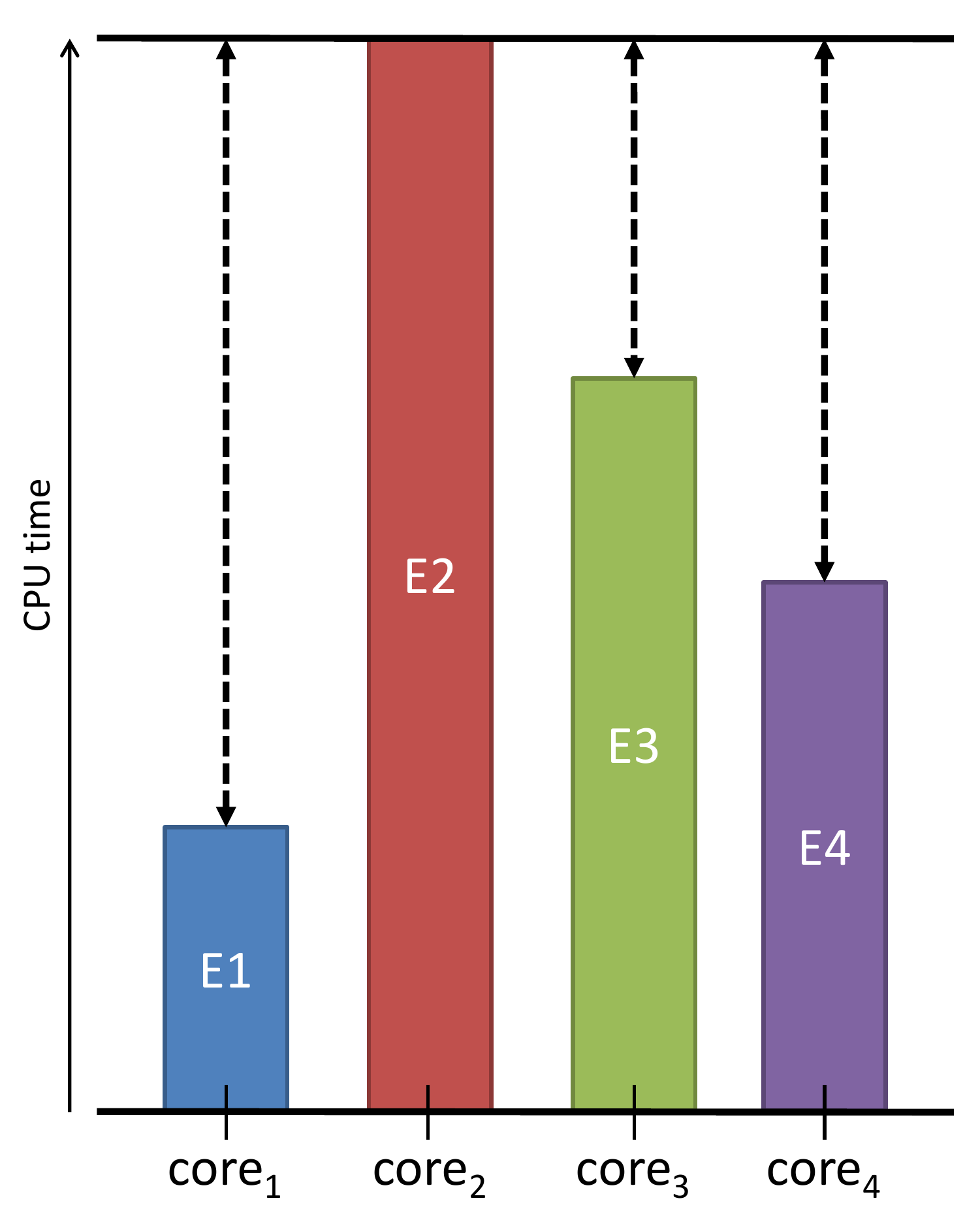}
\subcaption{Expert-wise}\label{fig:expert-wise}
\end{minipage}
\caption{Two options for parallelizing expert optimization in FAB M-step.}\label{fig:parallelize_options}
\end{figure}

%As we see in Fig.~\ref{fig:fab_proctime_ratio}, 
%The most significant computation bottleneck in learning FAB/HMEs is the expert optimization in the FAB M-step.
This subsection discusses advantages of sample-wise parallelization~(i.e., Algorithm~\ref{alg:comp}) over expert-wise parallelization.
Although the later one might seem to be a natural choice for parallelizing the FAB/HME learning algorithm, since expert optimizations can be processed independently, we observe two issues.
First, as Fig.~\ref{fig:expert_opt_time} shows, the distribution of processing time for expert optimization is heavy tailed.
The optimization with long process time has the large number of assigned instances, $N_{\phi_{j}}^{(t-1)}$, which 
varies across experts through the EM iterations.
%
%
%strongly affected by the number of features to be selected,
%
%\noindent{\textit{Unbalanced loads of parallelized tasks:}}
%Processing time for expert optimization is strongly affected by the number of features to be selected,
%which is highly correlated to the expected number of assigned instances, $N_{\phi_{j}}^{(t-1)}$, which 
%varies across experts through the EM iterations.
%Fig.~\ref{fig:expert_opt_time} shows a histogram of running time for the expert optimization.
%In most of the cases, computational times range from less than 1 second to 8 seconds, and the average is 2.4 seconds.
This implies (and we have empirically observed) that the loads of CPU cores get unbalanced, and fast ones have to wait 
for slow ones, as shown in Fig.~\ref{fig:expert-wise}, resulting in poor CPU usage.
Second, 
%\noindent{\textit{Degree of parallelization:}}
The number of CPU cores that expert-wise parallelization can use is bounded by the number of experts.
In the FAB/HME learning procedure, irrelevant experts are automatically eliminated, and the number of experts decreases through the iterations.
Fig.~\ref{fig:history_num_experts} illustrates an example of decreasing numbers of active experts over FAB EM iterations on a simple simulation dataset.
In this experiment, we used a machine with 16 CPU cores. 
The algorithm utilized only half CPU cores, however, after the number of active experts became 8 at the $20$-th step.

Fig.~\ref{fig:sample-wise} illustrates our sample-wise parallelization of expert optimization.
Our sample-wise parallelization addresses both of the issues.
For the unbalance of loads across the experts, the time complexity of expert optimization is approximately linear w.r.t. sample size,
and hence the sample-wise parallelization can equally distribute computational loads.
In addition, for degree of parallelization, since each task processes optimization for all experts, 
shrinkage decrease in the number of experts accelerates overall computation rather than decreasing CPU usage.
%One potential disadvantage to increasing the number of data partitions is an increase in the overhead of data communications between the dedicated server and worker nodes.
%However, the risk of significant overhead is sufficiently-low because \dFAB\ and its RDD are so well-designed that the cost of data shuffling is lower than the computation cost of expert optimization.

\subsection{Practical Implementation Tips}
%{\noindent \textbf{Leverage Spark's Advanced RDD Management: }}
All the distributed operations of \dFAB\ are implemented as chains of RDD transformations.
The \dFABimpl\ first converts the training data into an RDD and repeatedly transforms one RDD into another RDD to calculate the variational distribution $q$ and the model parameters $\theta$.
A driver process invokes the $\star$ processes of \dFAB\ on executors by means of such RDD transformers as \texttt{map} and \texttt{reduce}, and the calculation results are collected in the driver process via \texttt{collect} or \texttt{count} functions.
Since all the distributed operations are RDD transformations, \dFABimpl\ is able to obtain the benefits of Spark, e.g., sophisticated task/job scheduling~\cite{delaysched} and RDD fault tolerance.

%{\noindent \textbf{Reduce Communication Cost with Closure: }}
A driver process and executors exchange the model parameters $\theta$ via dynamically-generated closures.
A driver distributes $\theta$ by generating a closure passed to a transformation function of Spark.
The \dFABimpl\ implements the logic of \dFAB\ as first-class parameterized functions.
By passing $\theta$ as a parameter of the parameterized function, the function generates another closure which calculates its logic with parameter $\theta$.
The \dFABimpl\ passes the closure, derived from $\theta$ to transformation functions, that executes \dFAB\ logic on executors.

Checkpointing of intermediate RDDs is important w.r.t. the correct execution and the performance improvement.
\dFAB\ is a kind of EM algorithm, which sometimes requires long iterations.
The \dFABimpl\ executes the EM iterations by chaining RDD operations.
A long chain of RDD operations increases the stack size of remote procedure calls and induces stack overflow at worst because Spark has to maintain the lineage of the operations for keeping fault tolerance.
Furthermore, it is likely to cause garbage collection for intermediately-generated elements of an RDD and to result in a need for their re-computation.
To deal with the issues, the \dFABimpl\ makes the checkpoints at every 20 iterations, which number has been determined on the basis of empirical knowledge from our experiments.

\section{Experiments and Discussions}\label{sec:eval}

%To evaluate the accuracy and the performance of \dFAB\@\footnote{In this section \dFAB\ represents both \dFAB\ and \dFABimpl\ and \sFAB\ represents both \sFAB\ and \sFABimpl\ for simplicity.}, we performed experiments on a Spark cluster.
We used Spark 1.6.0, and 8 worker nodes ran Spark executors on YARN~\cite{Vavilapalli:2013:AHY:2523616.2523633}.
Each server and worker node employed two Intel Xeon E5-2640 v3 processors (16 physical cores), a 256 GB memory, and a 1 TB 7.2K RPM HDD, and they were connected via a 1 Gbps Ethernet.
Observed and target variables were standardized in advance.
%{\rc The default number of partitions $|\mathcal{W}|$ was 128.}

\subsection{Comparison with the Serial Algorithm}

\begin{figure}[t]
\centering
\includegraphics[width=.45\linewidth]{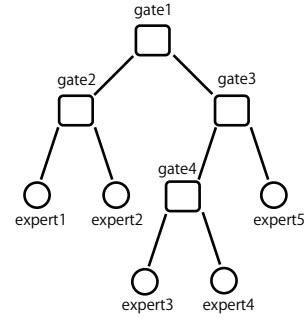}
\caption{The true tree structure of the artificial data.}\label{fig:simulation_model}
\end{figure}

We first demonstrate that \dFAB\ improves its execution speed with no negative impact on accuracy by comparing it with the serial algorithm, which is abbreviated as \sFAB, using an artificial regression data set used in the original FAB/HME paper~\cite{fab_eto14}.
% which was generated by the code used in the original FAB/HME paper~\cite{fab_eto14}.
The true tree structure illustrated in Fig.~\ref{fig:simulation_model} has 5 experts, and each expert uses 10--20 features.
On the $i$-th gating node, $g_i$ was fixed to 1, $\gamma_i$ and $t_i$ were randomly selected from $[1,D]$ and $[0,1]$, respectively.
On the $j$-th expert, the non-zero elements of $w_j$ were randomly sampled from $[0, 1]$.
%For regression test, 
$x^N$ and $y^N$ were sampled from $\text{Uniform}[0, 1]$ and $\mathcal{N}(y^{(n)}| w_j^T x^{(n)}, 0.1)$, where $N=1,000,000$, $5,000,000$, $10,000,000$ and $D=100$.
We employed $\delta = 5 \times 10^{-9}FIC^{(t-1)}$ (termination condition) and $\epsilon = 3 \times 10^{-2}N$ (shrinkage threshold).
The number of initial experts was $32$~(5-depth symmetric tree).

\begin{figure}[t]
\centering
\includegraphics[keepaspectratio, width=.95\linewidth]{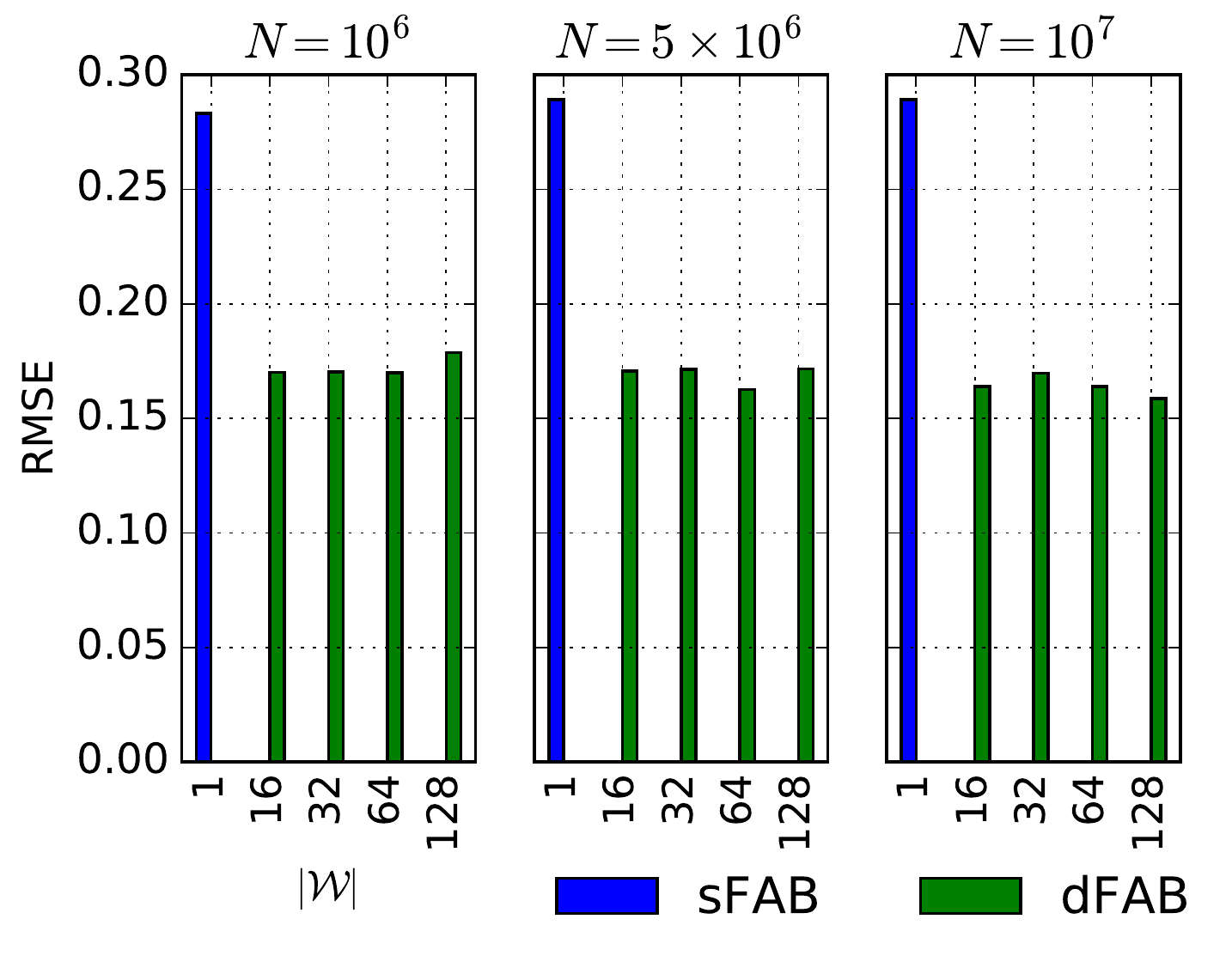}
\caption{Accuracy comparison between \dFAB\ and \sFAB\@.}\label{fig:artificial_acc}
\end{figure}

Fig.~\ref{fig:artificial_acc} illustrates the prediction accuracy of \dFAB\ and \sFAB\ over $|\mathcal{W}|$.
We observed:
\begin{itemize}[noitemsep,leftmargin=*]
\item The RMSE values of \dFAB\ were smaller than those of \sFAB\@. This might have been caused by the sample-wise parallelization which reduced the training variance as the original paper of the MESSAGE algorithm has also reported~\cite{NIPS2014_5328}.
\item The RMSE of \dFAB\ slightly increased with an increasing $|\mathcal{W}|$ with relatively small $N$. This might be because the number of samples on each executor becomes insufficient. %The reason of this is that increasing $|\mathcal{W}|$ decreased the number of selected features because $\frac{|\mathcal{W}|}{2}$ in \eqref{eq: voting} increases,
%and hence the final aggregated (voted and averaged) model employs a smaller number of features. 
However, the difference in accuracy became negligible with larger $N$, which was over 5,000,000 in this experimental condition.
\end{itemize}

\begin{figure*}[t]
\centering
\begin{minipage}[t]{.49\linewidth}
\centering
\includegraphics[keepaspectratio, width=\linewidth]{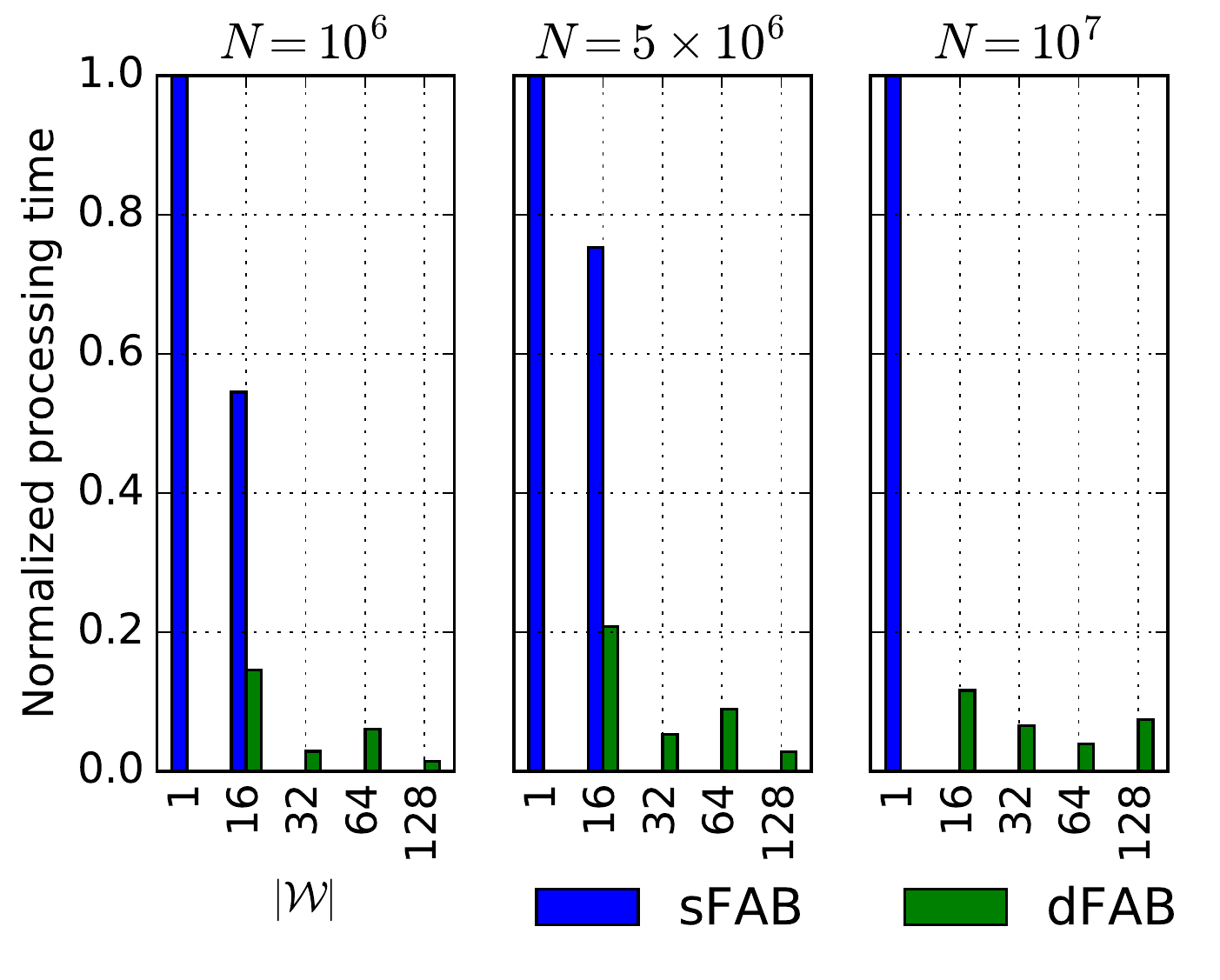}
\subcaption{Processing time until convergence.}\label{fig:artificial_totaltime_rg}
\end{minipage}
\begin{minipage}[t]{.49\linewidth}
\centering
\includegraphics[keepaspectratio, width=\linewidth]{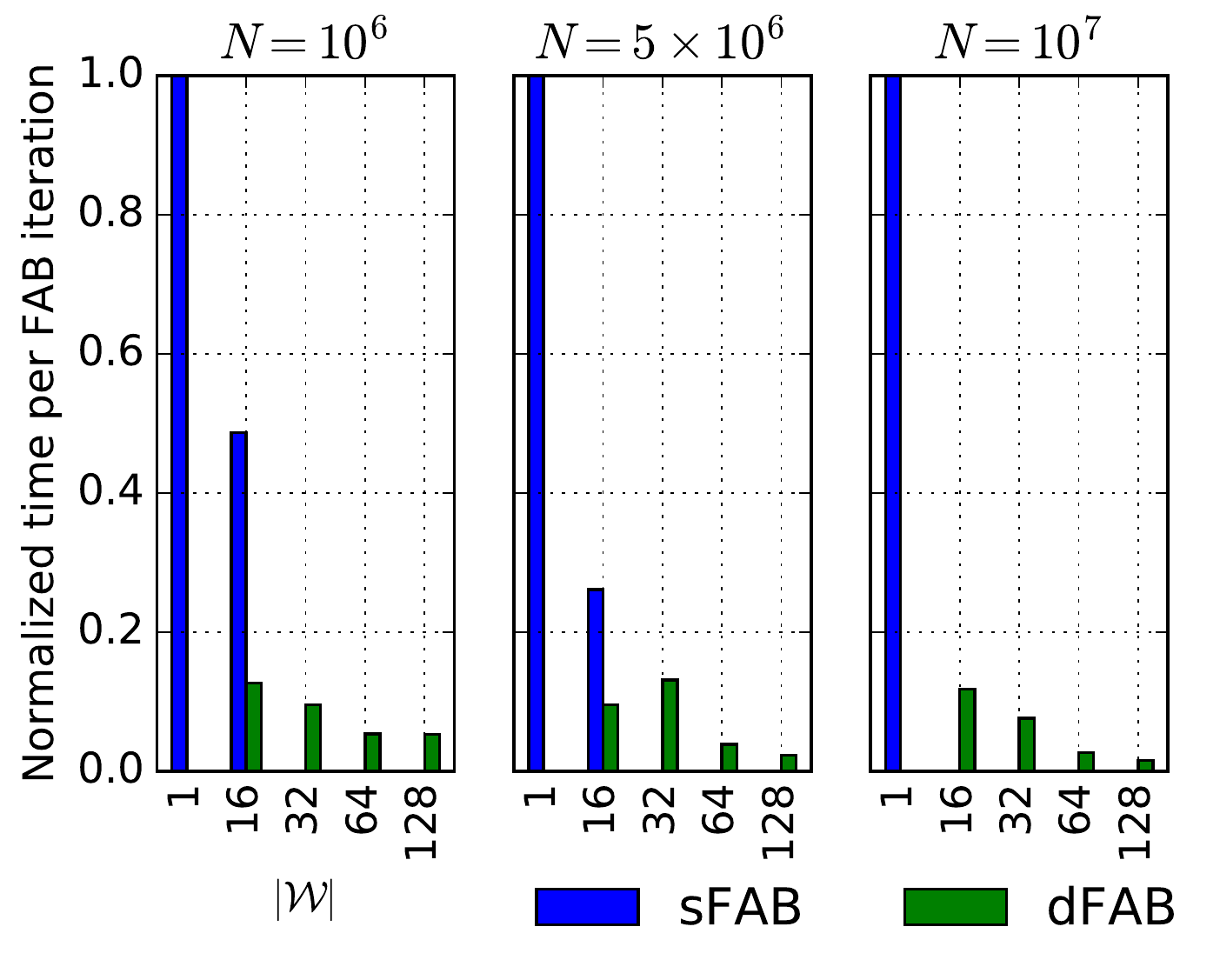}
\subcaption{Time per FAB EM iteration.}\label{fig:artificial_itertime_rg}
\end{minipage}
\caption{Processing times of \dFAB\ and \sFAB\@. The Y-axis represents normalized time by that of \sFAB\ with 1 CPU core.}\label{fig:artificial_proctime}
\end{figure*}

Fig.~\ref{fig:artificial_proctime} shows the processing times of \dFAB\ and \sFAB\ over $|\mathcal{W}|$ (i.e., number of CPU cores).
We implemented parallel gate optimization and expert-wise parallelization in \sFAB\ by OpenMP~\cite{openmp}. %, which corresponds to \sFAB\ where $|\mathcal{W}| = 16$ in Fig.~\ref{fig:artificial_proctime}.
From these results, we observed that:
\begin{itemize}[noitemsep,leftmargin=*]
\item %reduced its execution time in sub-linear related to the number of CPU cores; both the time per EM iteration and the total processing time until convergence were significantly lower than those for \sFAB\@. As discussed in Section~\ref{sec:sample-wise-par}, sample-wise parallelization of \dFAB\ outperformed expert-wise parallelization of \sFAB\@.
While EM iteration time for \sFAB\ with 16 CPU cores was as much as 73.8\% less than that with 1 CPU core, that for \dFAB\ with 16 CPU cores was as much as 90.4\% less than the same\footnote{Due to the memory limitation, \sFAB\ where $|\mathcal{W}| = 16$ could not finish where $N = 10,000,000$.}.
\item \dFAB\ with 128 CPU cores (i.e., $|\mathcal{W}| = 128$) showed an average time per EM iteration and a total processing time as much as 98.4\% and 98.5\% less than those for \sFAB\ with 1 CPU core, respectively. In general, larger $N$ achieves a better improvement of time per EM iteration with large $|W|$. %For example, the reduction ratios where $|\mathcal{W}| = 128$ were 94.7\% for $N = 1,000,000$, 97.7\% for $N = 5,000,000$ and 98.4\% for $N = 10,000,000$, respectively. 
A larger $N$ leads to a longer CPU processing time for each worker and, as a result, diminishes overhead in Spark execution, such as initial setup overhead for executors and RDDs and I/O wait for intermediate RDD checkpointing.
\item There existed a limitation in performance improvement over $|\mathcal{W}|$ depending on the value of $N$.
When $N = 1,000,000$, the performance improvement for EM iteration where $|\mathcal{W}| = 128$ was worse than that where $|\mathcal{W}| = 64$.
On the other hand, when $N = 5,000,000$ and $10,000,000$, the performance improvement for EM iteration where $|\mathcal{W}| = 128$ was greater than that where $|\mathcal{W}| = 64$. This implies we should determine an adequate number of partitions, i.e., $|\mathcal{W}|$, on the basis of the size of training data.
%\item There existed abnormal degradation of the performance improvement, e.g., the processing time until convergence where $|\mathcal{W}| = 64$ and $N = 1,000,000$ and the time per iteration where $|\mathcal{W}| = 32$ and $N = 5,000,000$. These unusual losses of the improvement are related to timings of checkpointing RDDs.
%The performance improvement ratio fluctuates based on how far the timing of the last checkpointing is from when \dFAB\ execution finishes.
%Despite that, the abnormality does not matter to the significant performance improvement of \dFAB\@.
\end{itemize}
%Note that \sFAB\ cannot run with $N=50,000,000$ due to a shortage of memory.

%Unlike \sFAB\@, \dFAB\ was able to finish its execution and successfully produce FAB/HMEs in 44,643 seconds.% and the test RMSE of the models was 0.262.
%The reason why the test RMSE became close to that of \sFAB\ is that the increase of $\frac{N}{|\mathcal{W}|}$ approximates the optimization results of $w \in \mathcal{W}$ at the expert optimization step.

\subsection{Benchmark Evaluation}

We compared the predictive accuracy of \dFAB\, on 3 regression (gas sensor array (CO), gas sensor array (methane) and household power consumption) and 2 classification data sets (HIGGS and HEPMASS) in UCI repository~\cite{uci}, with other distributed machine learning algorithms implemented in Spark MLlib.
The brief task abstracts of the data sets are described here;
\begin{itemize}[noitemsep,leftmargin=*]
\item The task of gas sensor array (CO) and (methane) data sets is multivariate regression with multiple responses based on chemical sensing system which consists of an array of 16 metal-oxide gas sensors and an external mechanical ventilator to simulate the biological respiration cycle.
\item The task of household power consumption data set is to predict the amount of power consumption every minutes of a house, which 
contains three meter's measurements of electric power consumption in one household with a one-minute sampling rate over a period of almost 4 years.
\item The task of HIGGS data set is a classification problem to distinguish between a signal process which produces Higgs bosons and a background process which does not, which contains 42 features in total. We used 27 features of all due to the limitation of the \dFAB\ implementation.
%The data has been produced using Monte Carlo simulations.
%The first 21 features are kinematic properties measured by the particle detectors in the accelerator.
%The last seven features are functions of the first 21 features; these are high-level features derived by physicists to help discriminate between the two classes.
\item The task of HEPMASS data set is to separate particle-producing collisions from a background source. 
%The search for exotic particles requires sorting through a large number of collisions to find the events of interest.
%This data set challenges one to detect a new particle of unknown mass.
This data set produced by Monte Carlo simulations contains 28 features. % (22 low-level features then 5 high-level features), and a 28th mass feature.
\end{itemize}
From Spark MLlib we chose ElasticNet~\cite{Zou05}, decision tree (DecTree), random forests (RF) for both classification and regression as baseline algorithms.
Note that RF is less interpretable and we evaluated it as a state-of-the-art distributed non-linear model. %While ElasticNet and DecTree are sort of highly-interpretable models, RF is one of models which tend to achieve high accuracy but are low interpretable.
We used 2-loop cross validation with 10-fold outer loops for evaluating test prediction error and 3-fold inner loops for parameter selection.
Note that FAB/HMEs do not need 2-loop cross validation, so \dFAB\ does not execute the inner loop.
We employed $\delta = 10^{-4}$ (termination condition) and $\epsilon=N \times 10^{-2}$ (shrinkage threshold). 
The number of initial experts was $8$ (3-depth symmetric tree).

\begin{table*}[t]
\centering
\caption{Test prediction errors (RMSEs for regression and 0-1 errors for classification). The numbers shown in parentheses are standard deviations. The best and second best methods are highlighted in \textbf{bold} and \textbf{\textit{bold italic}} faces, respectively.} \label{tbl:bench_err}
\begin{tabular}{c|c|c||c|c|c|c}
data&sample&dim&\dFAB&ElasticNet&DecTree&RF\\ \hline
\hline
gas sensor array (CO)&4,208,261&16&\textbf{0.542}(0.01)&0.597(0.00)&0.587(0.01)&\textbf{\textit{0.576}}(0.00)\\ \hline
gas sensor array (methane)&4,178,504&16&\textbf{0.548}(0.00)&0.600(0.00)&0.561(0.00)&\textbf{\textit{0.553}}(0.00)\\ \hline
household power consumption&2,075,259&3&\textbf{0.524}(0.01)&0.531(0.00)&\textbf{\textit{0.529}}(0.00)&0.655(0.00)\\ \hline
HIGGS&11,000,000&27&\textbf{\textit{0.335}}(0.01)&0.358(0.00)&0.337(0.00)&\textbf{0.317}(0.00)\\ \hline
HEPMASS&7,000,000&28&\textbf{0.156}(0.00)&\textbf{\textit{0.163}}(0.00)&0.167(0.00)&0.175(0.00)\\ \hline
\end{tabular}
\end{table*}
Table~\ref{tbl:bench_err} summarizes test RMSEs for regression data sets and classification errors for classification data sets.
For all three regression data sets, \dFAB\ outperformed the others while RF performed better than \dFAB\ on the HIGGS data for classification.
For the HIGGS data set, \dFAB\ generated FAB/HMEs with 2-5 active experts whose cardinalities were 5-14, that were much more interpretable than the models learned by RF, which consists of 300 trees with 5-depth.
%For example, on a HIGGS data set, GBTs generated 20 trees which have 32 leaf nodes in their models but \dFAB\ generated FAB/HMEs with 2--5 active experts whose cardinalities were 5--14.
In summary, the results indicate that 1) \dFAB\ achieved better predictive accuracy than that of other distributed algorithms implemented in Spark MLlib, 
and 2) it achieved competitive accuracy with more interpretable models than non-linear models of Spark MLlib.

\begin{figure*}[t]
\centering
\begin{minipage}[t]{.32\linewidth}
\centering
\includegraphics[keepaspectratio, width=.95\linewidth]{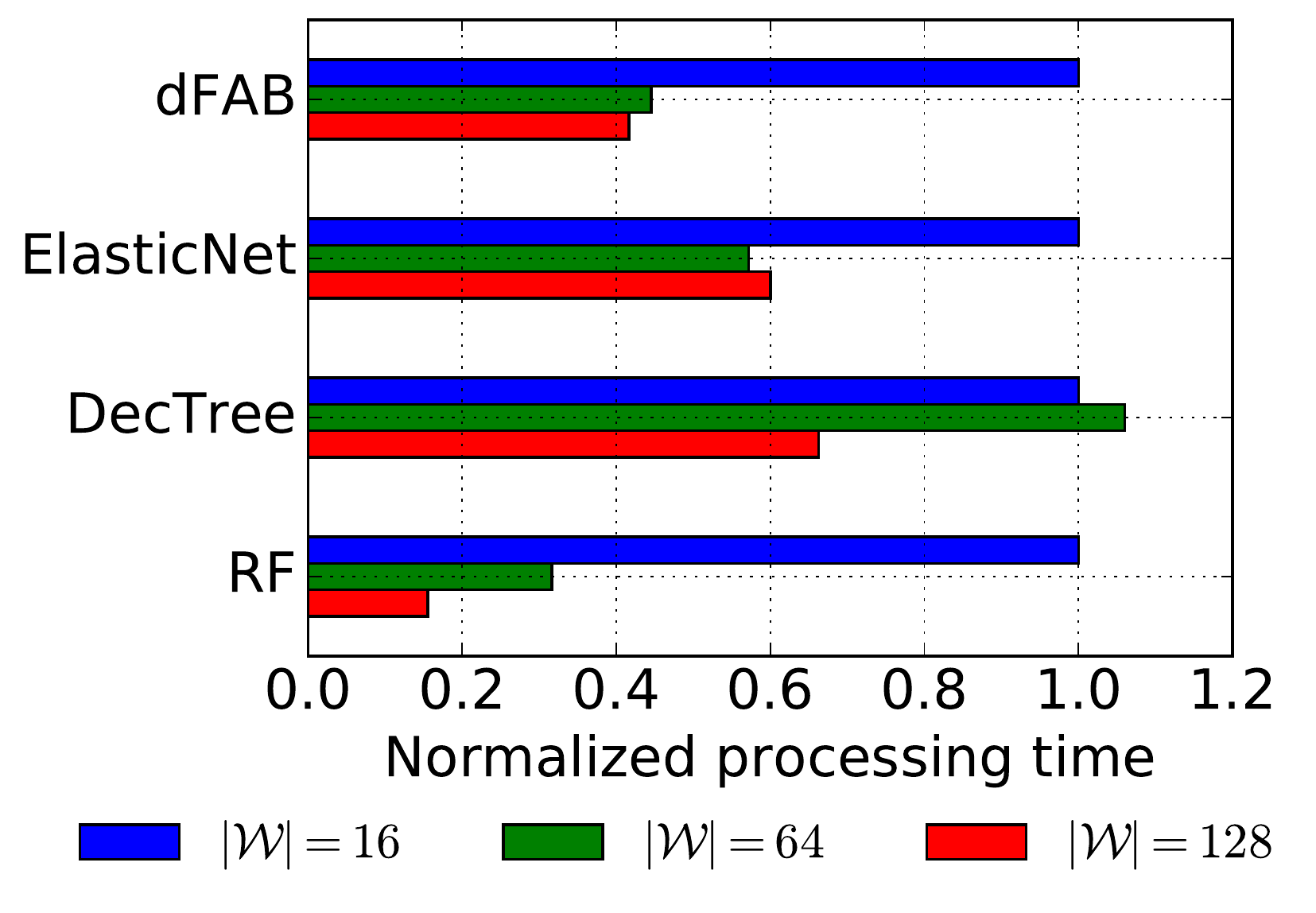}
\subcaption{gas sensor array (CO).}
\end{minipage}
\begin{minipage}[t]{.32\linewidth}
\centering
\includegraphics[keepaspectratio, width=.95\linewidth]{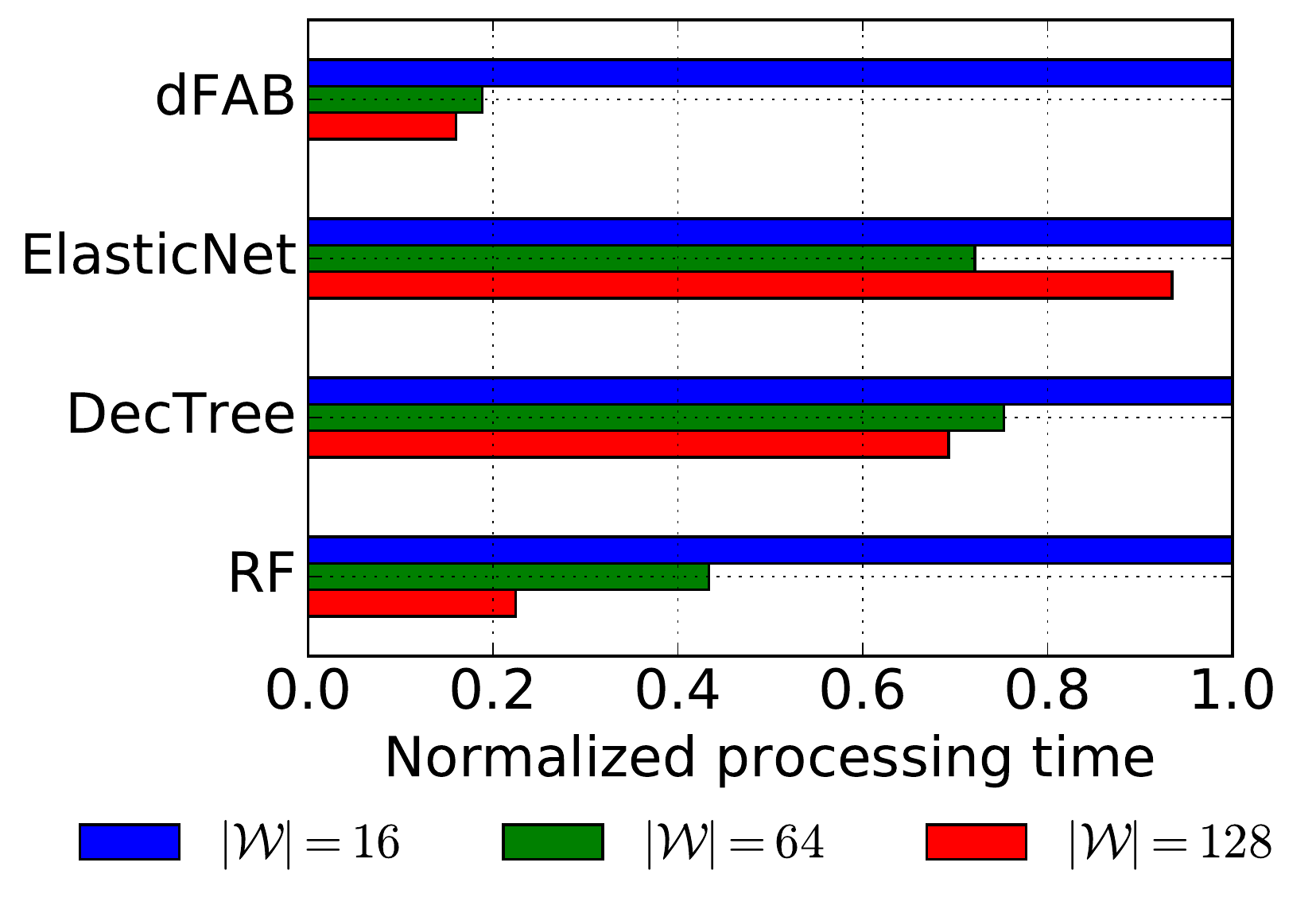}
\subcaption{gas sensor array (methane).}
\end{minipage}
\begin{minipage}[t]{.32\linewidth}
\centering
\includegraphics[keepaspectratio, width=.95\linewidth]{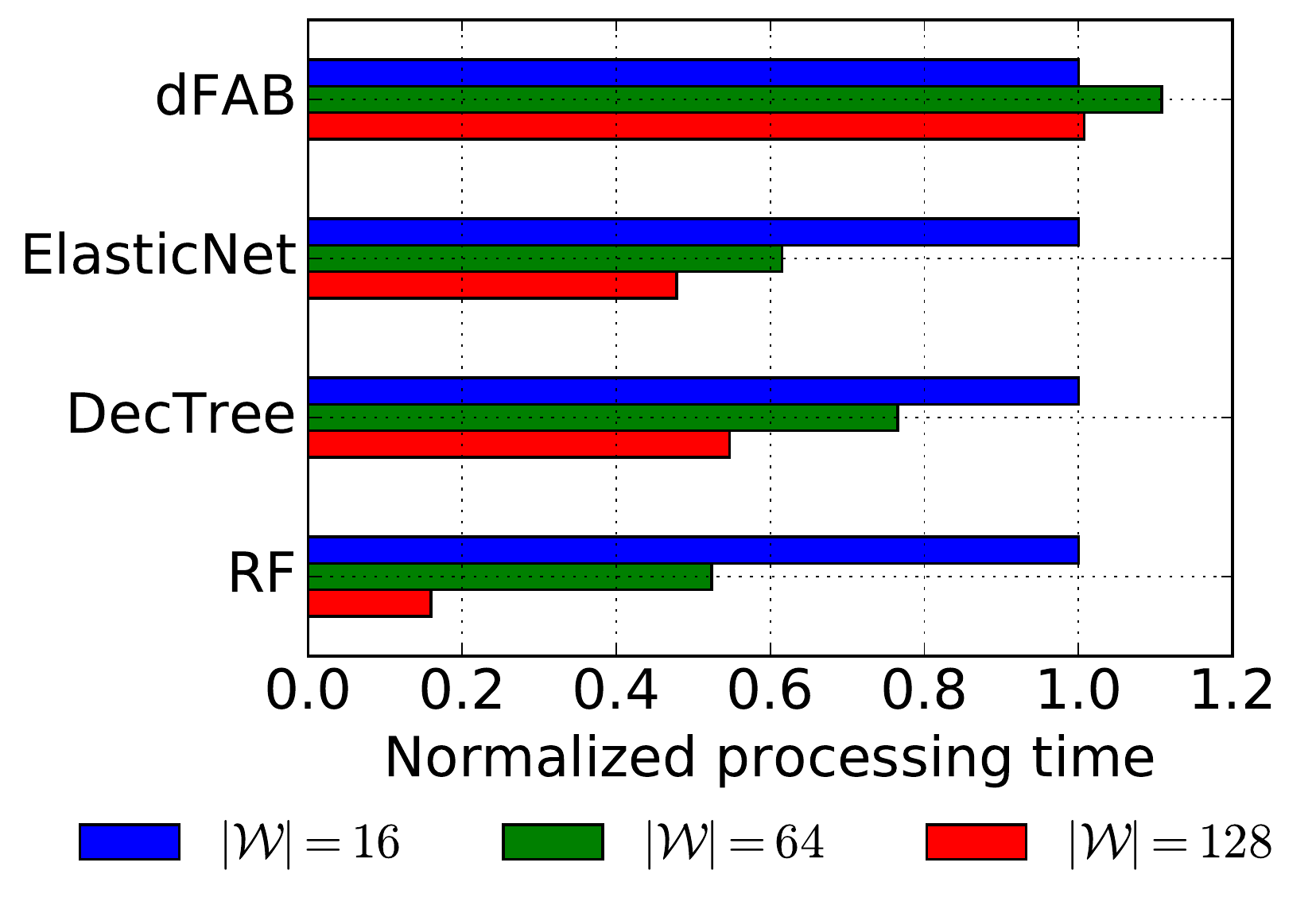}
\subcaption{household power consumption.}
\end{minipage}\\
\begin{minipage}[t]{.32\linewidth}
\centering
\includegraphics[keepaspectratio, width=.95\linewidth]{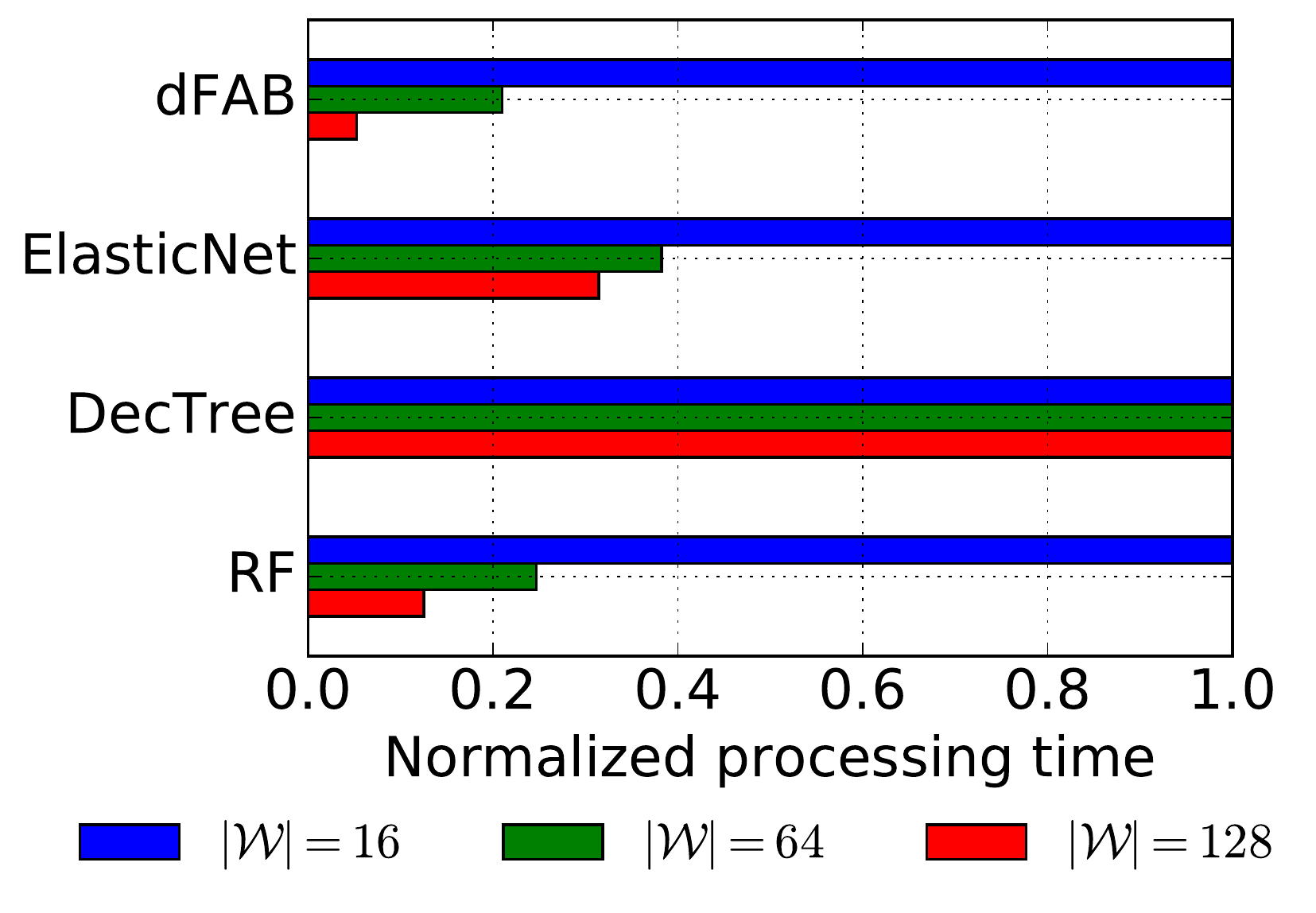}
\subcaption{HIGGS.}
\end{minipage}
\begin{minipage}[t]{.32\linewidth}
\centering
\includegraphics[keepaspectratio, width=.95\linewidth]{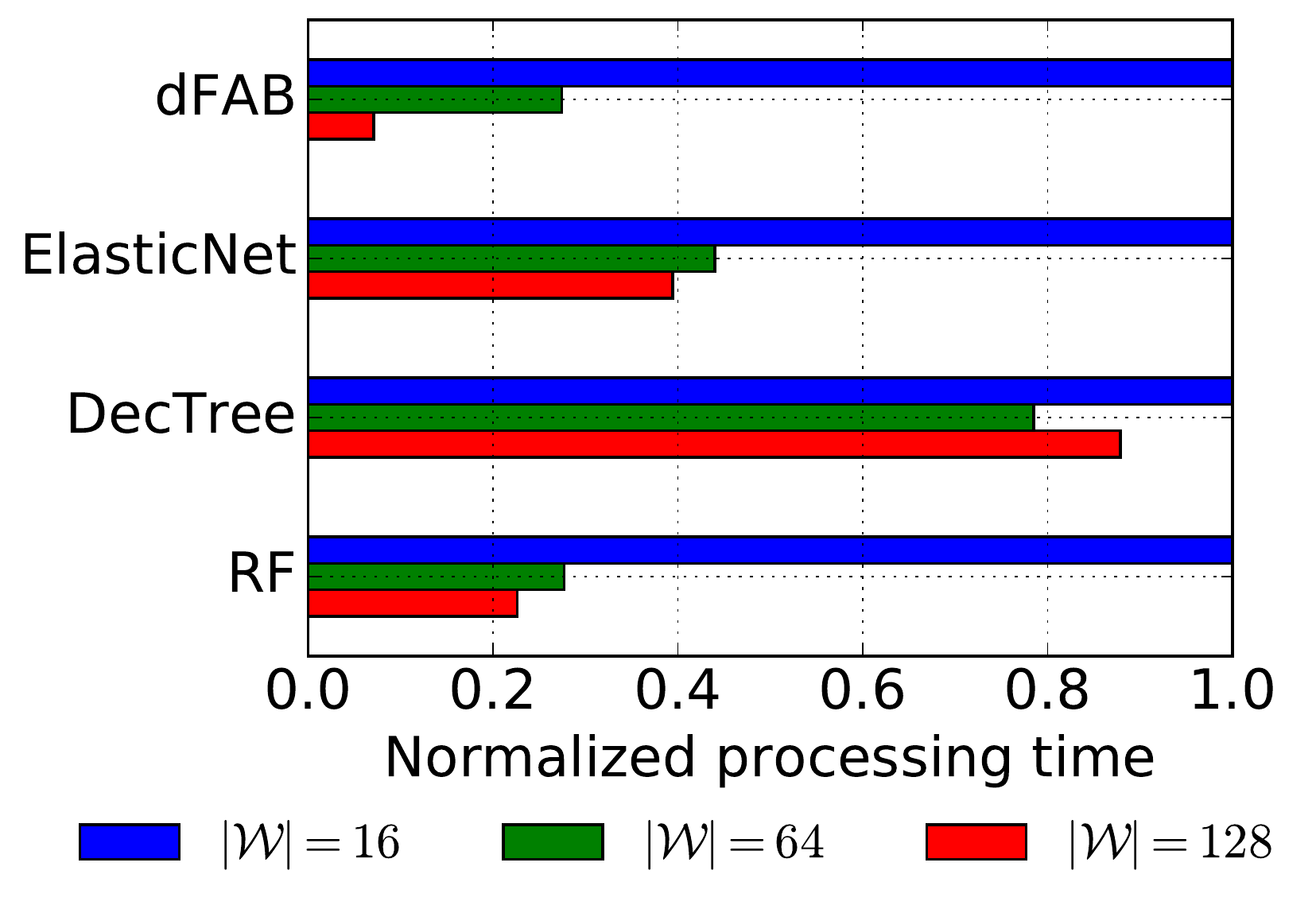}
\subcaption{HEPMASS.}
\end{minipage}
\caption{Comparison of the reduction of processing time between \dFAB\ and other algorithm implementations of Spark MLlib. Processing time is normalized by time of the same algorithm where $|\mathcal{W}| = 16$}\label{fig:proctime_opendata}
\end{figure*}

Next, %we demonstrate that the processing speed of \dFABimpl\ improves with number of CPU cores in the same way as other distributed computing implementation of machine learning algorithms.
we compared the reduction in processing time for \dFAB\ with that for other algorithm implementations of Spark MLlib over $|\mathcal{W}|$ as shown in Fig.~\ref{fig:proctime_opendata}. %compares the reduction ratio for processing times between \dFABimpl\ and other implementations of Spark MLlib over $|\mathcal{W}|$, i.e., a number of CPU cores.
We observed that:
\begin{itemize}[noitemsep,leftmargin=*]
\item \dFAB\ outperformed other implementations with gas sensor array (methane), HIGGS and HEPMASS\@.
\item With gas sensor array (CO), while \dFABimpl\ outperformed ElasticNet and DecTree, its improvement did not reach the level of RF\@.
The difference from gas sensor array (methane) data set was caused by the difference of convergence time derived from the difference of training data, even if the scale of data was similar to each other.
\item \dFAB\ had a limitation to scale out its computing performance on the household power consumption data.
This is because the data had only three features, which resulted in running the FAB EM optimization processes quite fast, and the overhead of Spark computation became a bottleneck.
\end{itemize}
In summary, these results reveal the performance scalability of \dFAB\ against state-of-the-art machine learning implementation on Spark. %becomes effective with data which consists of around five million instances with tens of features and is well effective with data which consists of more than five million instances with more than 20 features, such as HIGGS and HEPMASS\@.
%Note that \dFAB\ achieved better predictive accuracy and interpretability than the other algorithms of Spark MLlib, as demonstrated in the accuracy test, and the gap in scalability was not so large as to reduce the benefits of \dFAB\@.

\begin{table*}[t]
\centering
\caption{Processing time (seconds) with $|\mathcal{W}| = 128$.} \label{tbl:proc_time}
\begin{tabular}{c|c|c||c|c|c|c}
data&sample&dim&\dFAB&ElasticNet&DecTree&RF\\ \hline
\hline
gas sensor array (CO)&4,208,261&16&1,145&1,845&55&361\\ \hline
gas sensor array (methane)&4,178,504&16&379&1,827&68&475\\ \hline
household power consumption&2,075,259&3&289&49&27&206\\ \hline
HIGGS&11,000,000&27&1,396&196&75&968\\ \hline
HEPMASS&7,000,000&28&1,307&117&74&973\\ \hline
\end{tabular}
\end{table*}
Finally, we compared the processing time for \dFAB\ with that for other algorithm implementations of Spark MLlib.
In these experiments we set $|\mathcal{W}| = 128$ for all algorithm implementations.
Note that all processing time except \dFAB\ includes 3-fold inner loop.
As shown in Table~\ref{tbl:proc_time}, \dFAB\ averagely takes longer time than other algorithms because \dFAB\ solves non-convex optimization problems to achieve both high interpretability and accuracy.
However, as shown in Fig.~\ref{fig:proctime_opendata}, the execution performance scalability of \dFAB\ is better than others.
This property of \dFAB\ alleviates the performance disadvantage of \dFAB\ because we can easily reduce the execution time of \dFAB\ by just adding more worker nodes, which is not an unthinkable solution in this cloud computing era.

\subsection{Detailed Analysis on Resource Utilization}

\begin{figure*}[t]
\centering
\begin{minipage}[t]{.49\linewidth}
\centering
\includegraphics[keepaspectratio, width=.9\linewidth]{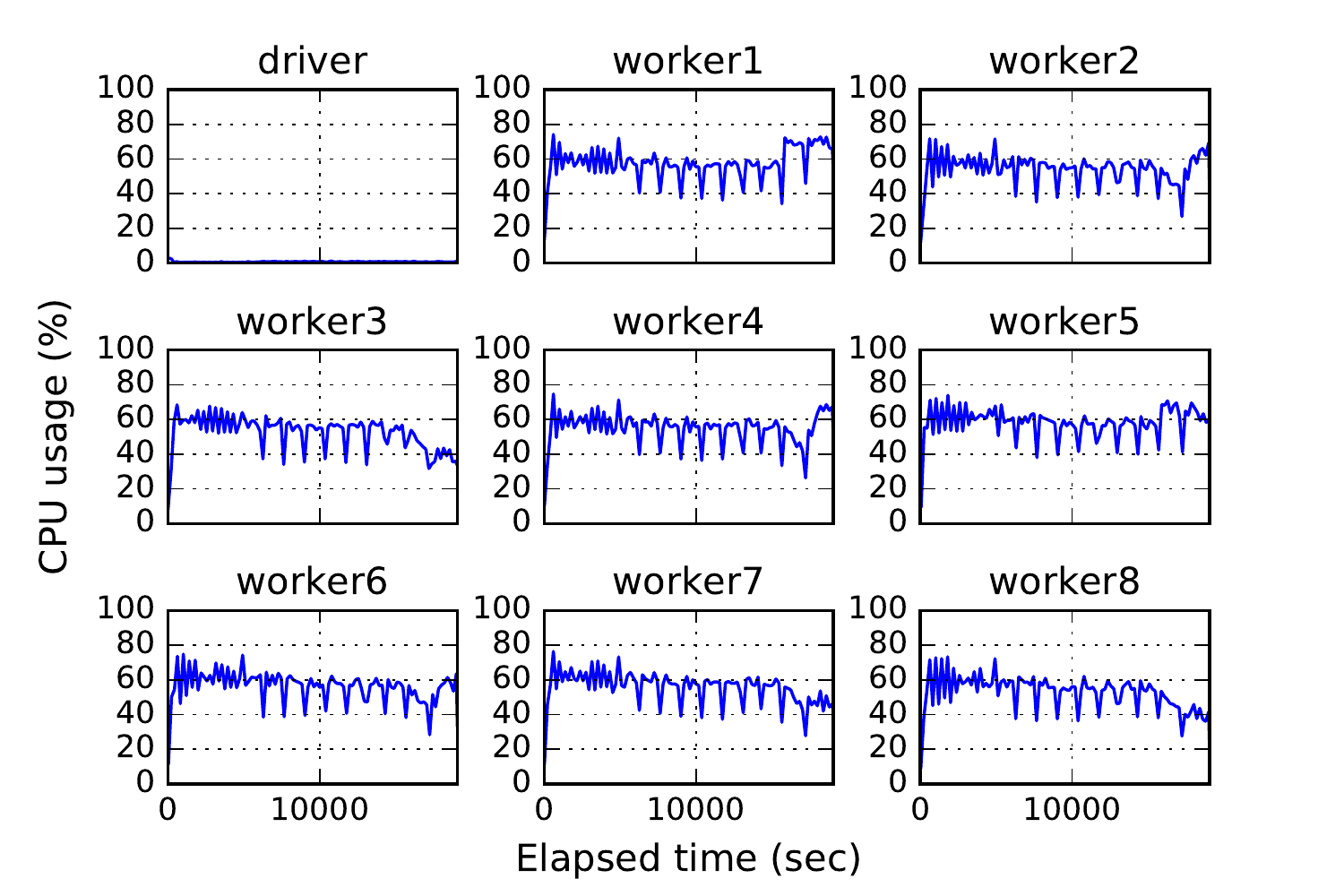}
\subcaption{CPU utilization.}
\end{minipage}
\begin{minipage}[t]{.49\linewidth}
\centering
\includegraphics[keepaspectratio, width=.9\linewidth]{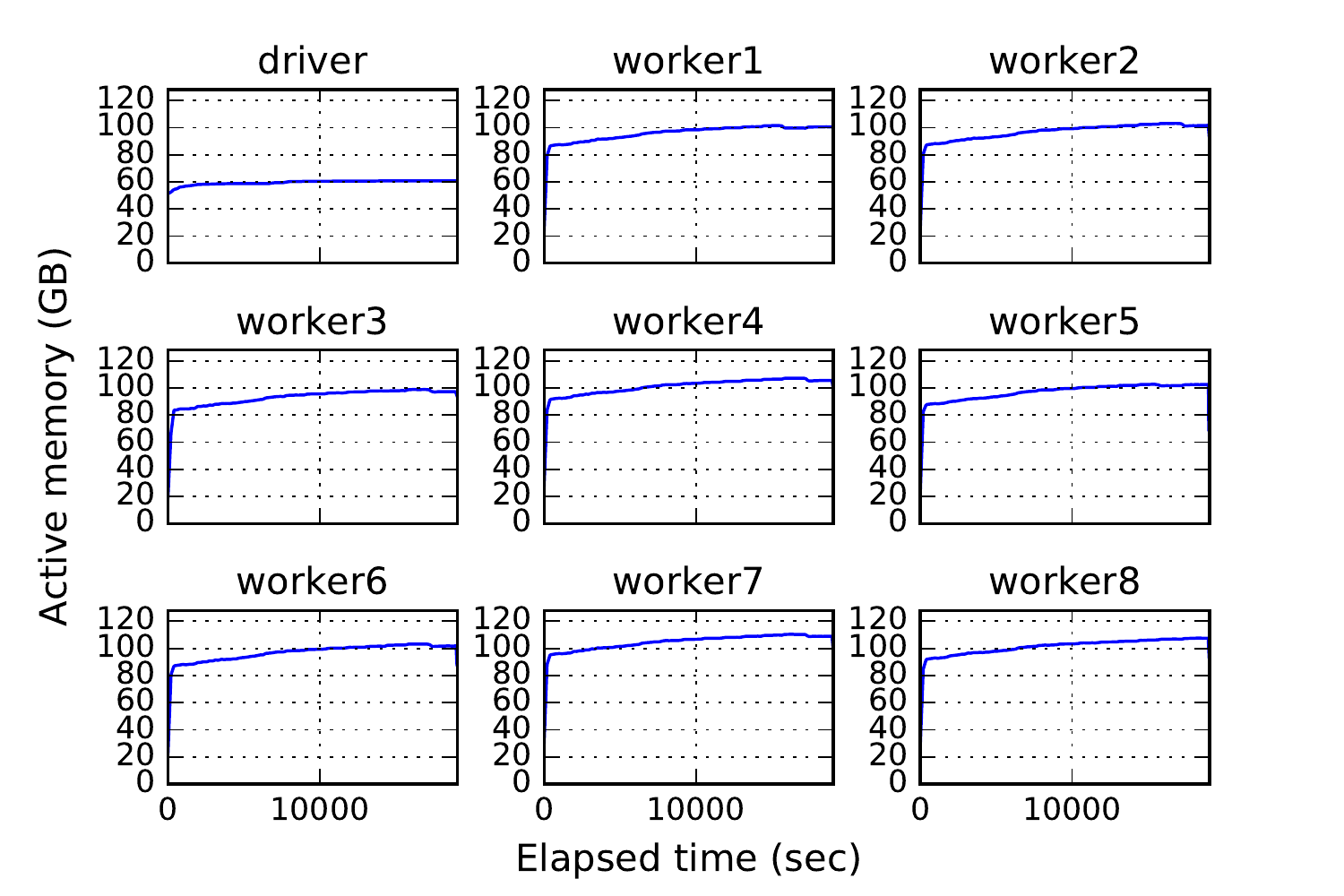}
\subcaption{Memory utilization.}
\end{minipage}\\
\begin{minipage}[t]{.49\linewidth}
\centering
\includegraphics[keepaspectratio, width=.9\linewidth]{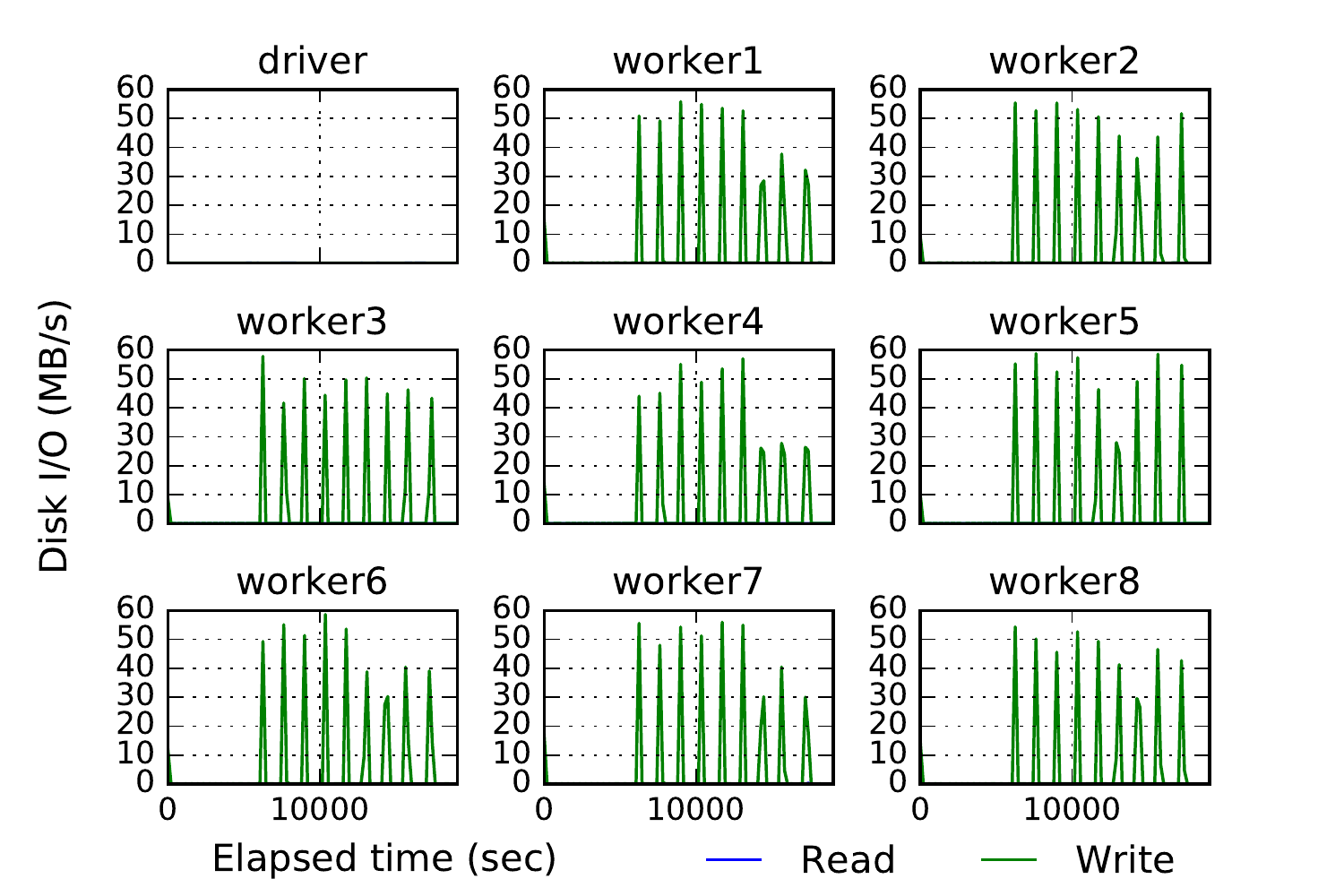}
\subcaption{Transfer rate of disk I/Os.}
\end{minipage}
\begin{minipage}[t]{.49\linewidth}
\centering
\includegraphics[keepaspectratio, width=.9\linewidth]{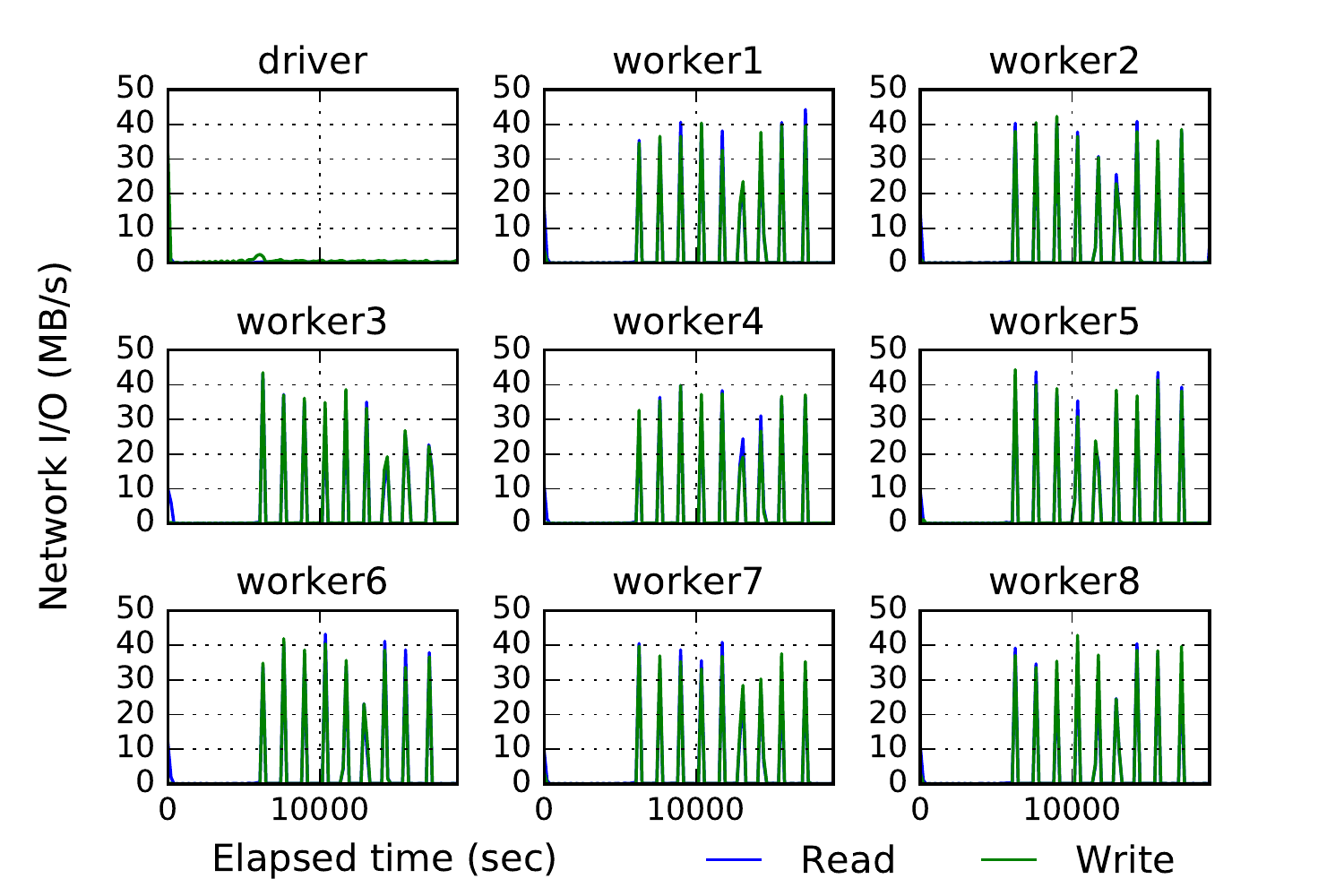}
\subcaption{Transfer rate of network I/Os.}
\end{minipage}
\caption{Resource consumption of \dFAB\@.}\label{fig:resource_util}
\end{figure*}

This section demonstrates that \dFAB\ can empirically utilize computing resources efficiently.
Fig.~\ref{fig:resource_util} illustrates the resource consumption of \dFAB\, which runs with the artificial data set  and where $N = 10,000,000$.
In the results, we observed:
\begin{itemize}[noitemsep,leftmargin=*]
\item CPU utilization of workers was roughly 60\% during the execution.
Some decrease in CPU utilization occurred due to the checkpointing RDDs. However, this did not severely degrade the performance of the execution.
\item The CPU core of the dedicated server for the driver process was mostly idle, and the process was not a bottleneck in the execution.
\item There were no spikes in memory utilization that implies \dFAB\ managed the memory utilization well.
\item Some spikes in disk and network access occurred due to the checkpointing RDDs on HDFS\@.
On the other hand, the average of disk and network accesses during the iterations without the checkpointing were less than several hundreds of KB/s.
\item The driver process did not issue notable disk accesses. This means that the execution of the driver was fully completed on memory and was not a performance bottleneck. Further, the driver process issued network traffic at the rate of tens of MB/s at the start of the execution.
This network transmission arose from the distribution process for the training data.
Once the process finished, the driver process transferred insignificant amounts of data with workers, which is less than hundreds of KB/s.
We can diminish the network transfer for training data by simple modification of the implementation to make the workers read the data from HDFS directly.
%\item Checkpointing RDDs decreased network communication and resulted in the reduction of the processing time as well.
%Fig.~\ref{fig:network_no-checkpoint} illustrates network communication with no checkpointing RDDs.
%The average amount of data transfer with and without checkpointing RDDs were {\rc XX.X} MB/s and {\rc YY.Y} MB/s, respectively.
%The processing time with no checkpointing RDDs exceeded that with checkpointing RDDs by {\rc ZZ.Z}\%.
\end{itemize}
%\begin{figure}
%\centering
%%\includegraphics[keepaspectratio, width=.9\linewidth]{figs/network_no-checkpoint}
%\caption{Network communication rate of \dFABimpl\ with no checkpointing RDDs.}\label{fig:network_no-checkpoint}
%\end{figure}
In summary, our results demonstrate that there are no serious performance bottlenecks in \dFAB\ and that efficient resource utilization resulted in good performance scalability in \dFAB\@.

\subsection{Scale-out Scalability Evaluation}

To evaluate the scale-out (or horizontal scaling) scalability of \dFAB\, we conducted experiments on a cluster with tens to hundreds virtual instances in Amazon Web Services (AWS)\@.
We prepared two scaled-out clusters which have a large number of worker nodes whose CPU has a small number of cores; one consists of \texttt{c3.large} instances (denoted as \texttt{c3.large} cluster) and another one consists of \texttt{r3.large} instances (denoted as \texttt{r3.large} cluster).
For comparison, we also prepared two scaled-up clusters which have a small number of worker nodes whose CPU has a large number of cores; one consists of \texttt{c3.8xlarge} instances (denoted as \texttt{c3.8xlarge} cluster) and another one consists of \texttt{r3.8xlarge} instances (denoted as \texttt{r3.8xlarge} cluster).
The hardware specification of them is described in Table~\ref{tbl:aws_spec}.
This evaluation used gas sensor array (methane) data set.
\begin{table*}[t]
\centering
\caption{Hardware specification of AWS instances.} \label{tbl:aws_spec}
\begin{tabular}{c||c|c|c|c|c|c}
Type&CPU cores&Memory (GiB)&Storage (GB)&Networking Perf.&Processor&Clock (GHz)\\ \hline
\hline
\texttt{c3.large}&2&3.75&2x16 SSD&Moderate&Intel Xeon E5-2680 v2&2.8\\ \hline
\texttt{c3.8xlarge}&32&60&2x320 SSD&10 Gb&Intel Xeon E5-2680 v2&2.8\\ \hline
\texttt{r3.large}&2&15.25&1x32 SSD&Moderate&Intel Xeon E5-2670 v2&2.5\\ \hline
\texttt{r3.8xlarge}&32&244&2x320 SSD&10 Gb&Intel Xeon E5-2670 v2&2.5\\ \hline
\end{tabular}
\end{table*}

\begin{figure*}[t]
\centering
\begin{minipage}[t]{.45\linewidth}
\centering
\includegraphics[keepaspectratio, width=.95\linewidth]{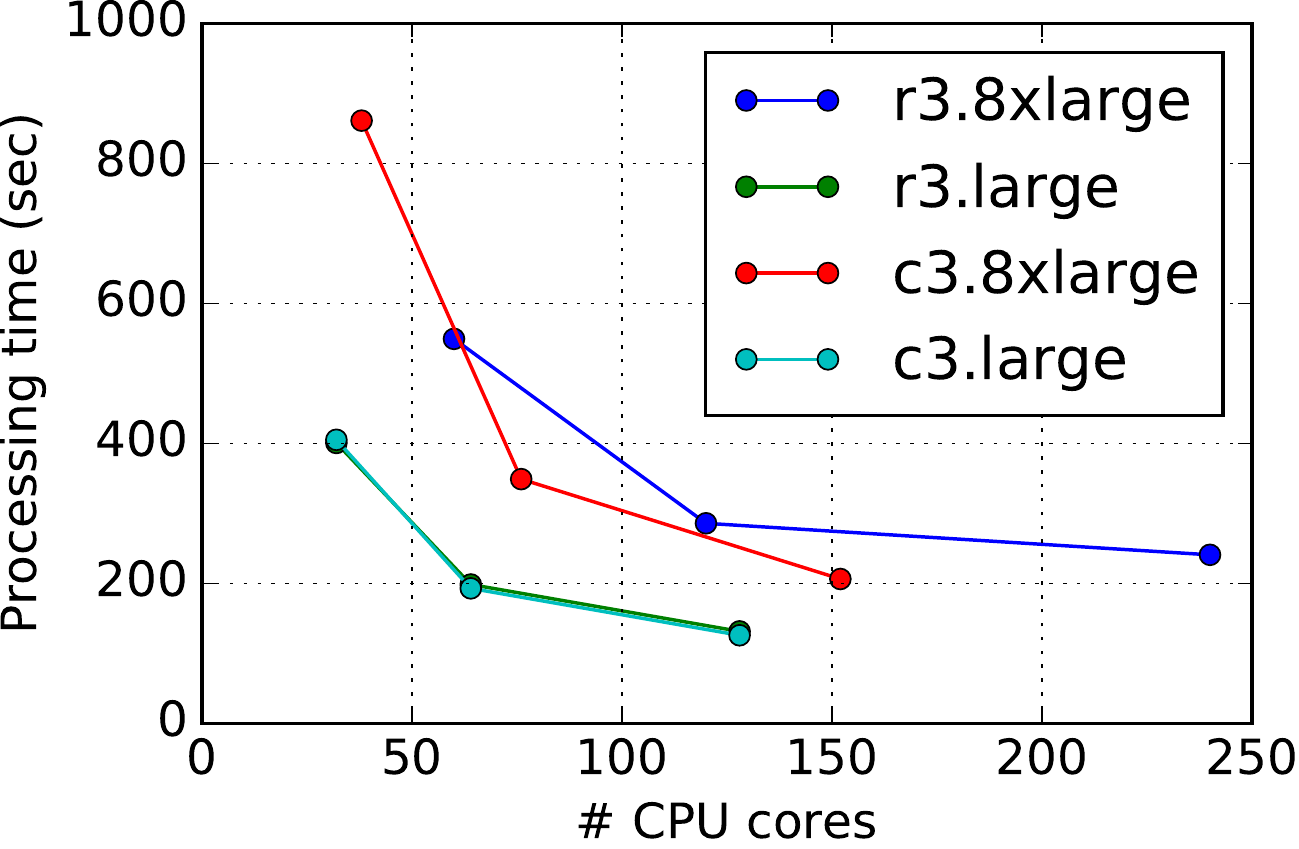}
\subcaption{Processing time until convergence.}\label{fig:totaltime_aws}
\end{minipage}
\begin{minipage}[t]{.45\linewidth}
\centering
\includegraphics[keepaspectratio, width=.95\linewidth]{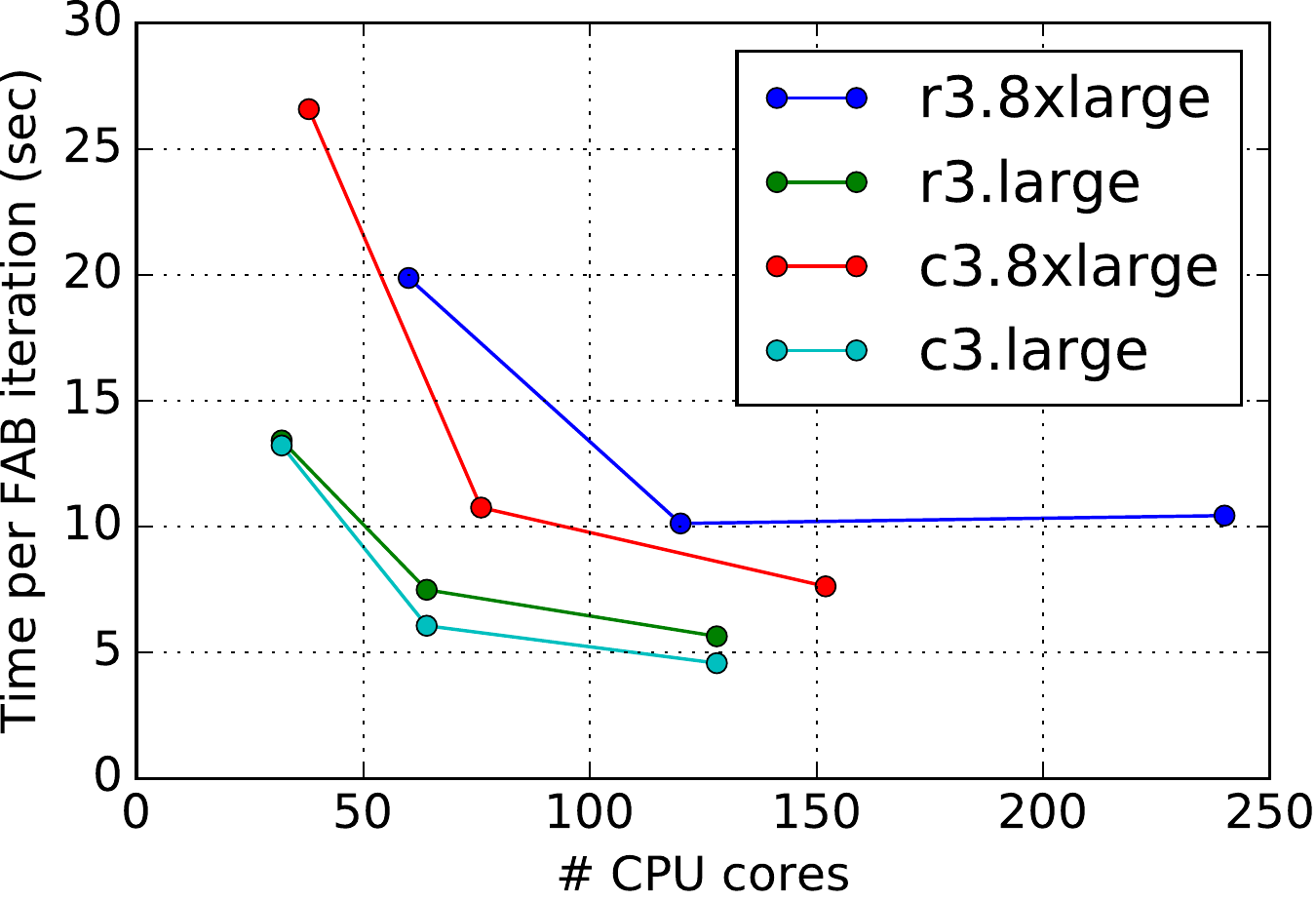}
\subcaption{Time per FAB EM iteration.}\label{fig:itertime_aws}
\end{minipage}\\
\caption{Processing time comparison between scaled-out clusters (\texttt{c3.large} and \texttt{r3.large}) and scaled-up clusters \texttt{c3.8xlarge} and \texttt{r3.8xlarge}).}\label{fig:proctime_aws}
\end{figure*}

Fig.~\ref{fig:totaltime_aws} and \ref{fig:itertime_aws} show average processing time and time per EM iteration of \dFAB\ on the clusters, respectively.
The experimental results show that the execution speed of \dFAB\ on \texttt{c3.large} and \texttt{r3.large} clusters (scaled-out clusters) is around 2 times faster than that of \texttt{c3.8xlarge} and \texttt{r3.8xlarge} clusters (scaled-up clusters)\footnote{Even with same total number of CPU cores, the execution time on \texttt{c3.large} and \texttt{r3.large} clusters was shorter than that on \texttt{c3.8xlarge} and \texttt{r3.8xlarge} clusters. This is because good scaled-out scalability of \dFAB\ exploits the performance advantages of the hardware architecture such as an automatically CPU clock-up technology and the large total bandwidth of memory access.}.
\dFAB\ achieved good scale-out scalability on \texttt{c3.large} and \texttt{r3.large} clusters despite scaled-out architectures tend to increase the network communications due to huge number of worker nodes.
This scale-out scalability is obtained by the design of \dFAB\ %and \dFABimpl\ 
which reduces the network traffic as much as possible to avoid performance degradation.

In summary, our experimental results demonstrate that the design of \dFAB\ algorithm and implementation possesses the capability of high performance scalability on a scaled-out cluster.

\section{Conclusions}\label{sec:conc}

In this paper we have proposed the distributed FAB/HME algorithm scalable algorithm to learn interpretable and accurate piecewise sparse linear models from Big Data.
By taking advantages of FAB inference and the MESSAGE algorithm with a novel asymptotic bias correction of FIC, \dFAB\ realizes fully automated model selection with linearly scale-out capability over the data size.
Further, we presented a design of \dFAB\ on Spark, the rising distributed memory computation platform.
%It processes large amounts of statistical data, such as data matrices and variational distributions, with no data shuffling between worker nodes.
%This paper has also presented a design and implementation strategy for \dFAB.
%This strategy features two important design ideas.
Our RDD for \dFAB\ enables us to fully utilize CPU without needing to shuffle data during optimization processes.
Our experimental results have demonstrated that \dFAB\ achieves high prediction accuracy and performance scalability for both synthetic and public data.
One of important future work is to explore the possibility of rebuilding the \dFAB\ algorithm so as to make it asynchronous, in order to  improve the efficiency of computing resource utilization and accelerate algorithm execution.

\bibliographystyle{IEEEtran}
\bibliography{ms}

\end{document}